\documentclass{article}

  \PassOptionsToPackage{numbers,sort,compress}{natbib}
  \usepackage[preprint]{nips_2018mod} 
\usepackage[utf8]{inputenc} % allow utf-8 input
\usepackage[T1]{fontenc}    % use 8-bit T1 fonts
\usepackage{hyperref}       % hyperlinks
\usepackage{url}            % simple URL typesetting
\usepackage{booktabs}       % professional-quality tables
\usepackage{amsfonts}       % blackboard math symbols
\usepackage{nicefrac}       % compact symbols for 1/2, etc.
\usepackage{microtype}   
  
\usepackage{here} 
\usepackage{amsmath}
\usepackage{amssymb}
\usepackage{microtype}
\usepackage{graphicx}
\usepackage{subfigure}
\usepackage{booktabs}
\usepackage{bm}
\usepackage{amsthm}
\newtheorem{theorem}{Theorem}[section]

\usepackage{color}

\newcommand{\comk}[1]{\textcolor{black}{#1}}

\begin{document}
\title{Pathological spectra of the Fisher information metric and its variants in deep neural networks}

\author{
 Ryo Karakida  
     \thanks{
       Artificial Intelligence Research Center, National Institute of Advanced Industrial Science and Technology (AIST), Tokyo, Japan, E-mail: \texttt{karakida.ryo@aist.go.jp}}, \ \ 
        Shotaro Akaho \thanks{National Institute of Advanced Industrial Science and Technology (AIST), Ibaraki, Japan,  E-mail: \texttt{s.akaho@aist.go.jp}}, \ \  
       and Shun-ichi Amari 
       \thanks{RIKEN Center for Brain Science (CBS), Saitama, Japan,     E-mail: \texttt{amari@brain.riken.jp}}  }

\maketitle

\begin{abstract}
The Fisher information matrix (FIM) plays an essential role in statistics and machine learning as a Riemannian metric tensor or a component of the Hessian matrix of loss functions. Focusing on the FIM and its variants in deep neural networks (DNNs), we reveal their characteristic scale dependence on the network width, depth and sample size when the network has random weights  and is sufficiently wide.
This study covers two widely-used FIMs for regression with linear output and for classification with softmax output. Both FIMs asymptotically show pathological eigenvalue spectra in the sense that a small number of eigenvalues become large outliers depending the width or sample size while the others are much smaller.  It implies that the local shape of the parameter space or loss landscape is very sharp in a few specific directions while almost flat in the other directions. In particular, the softmax output disperses the outliers and makes a tail of the eigenvalue density spread from the bulk.
We also show that pathological spectra appear in other variants of FIMs: one is the neural tangent kernel; another is a metric for the input signal and feature space that arises from feedforward signal propagation. Thus, we provide a unified perspective on  the FIM and its variants that will lead to more quantitative understanding of learning in large-scale DNNs. 
\end{abstract}

\section{Introduction}
\label{Introduction}
Deep neural networks (DNNs) have outperformed many standard machine-learning methods in practical applications \cite{nature2015}. Despite their practical success, many theoretical aspects of DNNs remain to be uncovered, and there are still many heuristics used in deep learning. We need a solid theoretical foundation for elucidating how and under what conditions DNNs and their learning algorithms work well. 

The Fisher information matrix (FIM) is a fundamental metric tensor that appears in statistics and machine learning. 
An empirical FIM is equivalent to the Hessian of the loss function around a certain global minimum, and it affects the performance of optimization in machine learning. In information geometry, the FIM defines the Riemannian metric tensor of the parameter manifold of a statistical model \cite{amari2016information}. The natural gradient method is a first-order gradient method in the Riemannian space where the FIM works as its Riemannian metric \cite{ama1998,park2000,pascanu2013,ollivier2015riemannian,martens2015optimizing}. 
The FIM also acts as a regularizer to prevent catastrophic forgetting \cite{kirkpatrick2017overcoming}; a DNN trained on one dataset can learn another dataset without forgetting information if the parameter change is regularized with a diagonal FIM. 

However, our understanding of the FIM for neural networks has so far been limited to empirical studies and theoretical analyses of simple networks. Numerical experiments empirically confirmed that the eigenvalue spectra of the FIM and those of the Hessian are highly distorted; that is, most eigenvalues are close to zero, while others take on large values \cite{lecun1998efficient,sagun2017empirical,papyan2019measurements,ghorbani2019investigation}. Focusing on shallow neural networks, \citet{pennington2018spectrum} theoretically analyzed the FIM's eigenvalue spectra by using random matrix theory, and \citet{fuk1996} derived a condition under which the FIM becomes singular. \citet{liang2017fisher} have connected FIMs to the generalization ability of DNNs by using model complexity, but their results are restricted to linear networks. Thus, theoretical evaluations of deeply nonlinear cases seem to be difficult mainly because of iterated nonlinear transformations. To go one step further, it would be helpful if a framework that is widely applicable to various DNNs could be constructed. 

Investigating DNNs with random weights has given promising results. When such DNNs are sufficiently wide, we can formulate their behavior by using simpler analytical equations through coarse-graining of the model parameters, as is discussed in mean field theory \cite{amari1974method,kadmon2016,poole2016,schoenholz2016,yang2017,xiao2018dynamical,yang2019scaling} and random matrix theory \cite{pennington2018emergence,pennington2017geometry,pennington2017nonlinear}. For example, \citet{schoenholz2016} proposed a mean field theory for backpropagation in fully-connected DNNs. This theory characterizes the amplitudes of gradients by using specific quantities, i.e., {\it order parameters} in statistical physics, and enables us to quantitatively predict parameter regions that can avoid vanishing or explosive gradients. This theory is applicable to a wide class of DNNs with various non-linear activation functions and depths. Such DNNs with random weights are substantially connected to Gaussian process and kernel methods \cite{daniely2016toward,lee2017deep,matthews2018gaussian,jacot2018neural}. Furthermore, the theory of the neural tangent kernel (NTK) explains that even trained parameters are close enough to the random initialization in sufficiently wide DNNs and the performance of trained DNNs is determined by the NTK on the initialization \cite{jacot2018neural,lee2019wide,arora2019exact}.

\citet{karakida2018universal} focused on the FIM corresponding to the mean square error (MSE) loss and proposed a framework to express certain eigenvalue statistic by using order parameters. They revealed that when fully-connected networks with random initialization are sufficiently wide, the FIM’s  eigenvalue spectrum asymptotically becomes pathologically distorted. As the network width increases,  a small number of the eigenvalues asymptotically take on huge values and become outliers while the others are much smaller. 
The distorted shape of the eigenvalue spectrum is consistent with empirical reports \cite{lecun1998efficient,sagun2017empirical,papyan2019measurements,ghorbani2019investigation}. 
While \citet{le1991eigenvalues} 
implied that such pathologically large eigenvalue might appear in multi-layered networks and affect the training dynamics, its theoretical elucidation has been limited to  a data covariance matrix in
a linear regression model. The results of \cite{karakida2018universal} can be regarded as a theoretical verification of this large eigenvalue  suggested by \cite{le1991eigenvalues}.
The obtained eigenvalue statistics have given insight into the convergence of gradient dynamics \cite{lecun1998efficient,karakida2018universal}, mechanism of batch normalization to decrease the sharpness of the loss function \cite{karakida2019normalization}, and generalization measure of DNNs based on the minimum description length \cite{Sun2019}. 

In this paper, we extend the framework of the previous work \cite{karakida2018universal} and reveal that various types of FIMs and variants show pathological spectra. Our main contribution is the following:
\begin{itemize}
\item {{\bf FIM for classification tasks with softmax output}: While the previous works \cite{karakida2018universal,karakida2019normalization} analyzed the FIM for regression based on  the MSE loss, we typically use the cross-entropy loss with softmax output in classification tasks.
We analyze this FIM for classification tasks and reveal that its spectrum is  pathologically distorted as well. While the FIM for regression tasks has unique and degenerated outliers in the infinite-width limit, the softmax output can make these outliers disperse and remove the degeneracy.
Our theory shows that there are number-of-classes outlier eigenvalues, which is consistent with experimental reports \cite{sagun2017empirical,papyan2019measurements}.
Experimental results demonstrate that the eigenvalue density has a tail of outliers spreading form the bulk. 
} 
\end{itemize}

Furthermore, we also give a unified perspective on the variants:  
\begin{itemize}
\item{{\bf Diagonal Blocks of FIM:} We give a detailed analysis of the diagonal block parts of the FIM for regression tasks. Natural gradient algorithms often use a block diagonal approximation of the FIM \cite{amari2018fisher}. We show that the diagonal blocks also suffer from pathological spectra. }
\item {{\bf Connection to NTK}: The NTK and FIM inherently share the same non-zero eigenvalues. Paying attention to a specific re-scaling of the parameters assumed in studies of NTK, we clarify that NTK's eigenvalue statistics become independent of the width scale. Instead, the gap between the average and maximum eigenvalues increases with the sample size. This suggests that, as the sample size increases, the training dynamics converge non-uniformly and that calculations with the NTK become ill-conditioned. We also demonstrate a simple 
normalization method to make eigenvalue statistics that are independent of both the width and the sample size.}
\item {{\bf Metric tensors for input and feature spaces}: We consider metric tensors for input and feature spaces spanned by neurons in input and hidden layers. These metric tensors potentially enable us to evaluate the robustness of DNNs against perturbations in the input and feedforward propagated signals. We show that the spectrum is pathologically distorted, similar to FIMs, in the sense that the outlier of the spectrum is much far from most of the eigenvalues. The softmax output makes the outliers disperse as well.}
\end{itemize}

In summary, this study sheds light into the asymptotical eigenvalue statistics common to various wide networks.

\section{Preliminaries}
\subsection{Model}
We investigated the fully-connected feedforward neural network shown in Fig. 1. The network consists of one input layer, $L-1$ hidden layers ($l=1,...,L-1$), and one output layer. It includes shallow nets ($L=2$) and arbitrary deep nets ($L\geq 3$). The network width is denoted by $M_l$. The pre-activations $u^l_i$ and activations of units $h_i^l$ in the $l$-th layer are defined recursively by
\begin{equation}
u_i^l= \sum_{j=1}^{M_{l-1}} W_{ij}^{l} h_j^{l-1} +b_i^l, \ \ h_i^{l}= \phi(u_i^{l}), \label{eq1}
\end{equation}
 which will be explained in the following. The input signals are $h_i^0=x_i$, which propagate layer by layer by Eq. (\ref{eq1}). We define the weight matrices as $W_{ij}^l \in \mathbb{R}^{M_l \times M_{l-1}}$ and the bias terms as $b_{i}^l \in \mathbb{R}^{M_l}$. Regarding the network width, we set
\begin{equation}
M_l = \alpha_l M \ \ ( l\leq L-1), \ \ M_L=C, 
\end{equation}
and consider the limiting case of a sufficiently large $M$ with constant coefficients $\alpha_l>0$. The number of output units is taken to be a constant $C$, as is usually done in practice. 
We denote the linear output of the last layer by
\begin{equation}
 f_i=u_i^L. \label{eq2:0323} 
\end{equation}
We also investigate DNNs with softmax outputs   in Section \ref{Sec_cross}. 
The $C$-dimensional softmax function is given by
\begin{equation}
 g_i := \frac{\exp(f_i)}{\sum_{k=1}^C \exp(f_k)}, \label{eq25:0731}
\end{equation}
for $i=1,...,C$.

FIM computations require the chain rule of backpropagated signals $\delta_k^l \in \mathbb{R}^{M_l}$. The backpropagated signals are defined by $\delta_{k,i}^{l} := \partial f_{k}/\partial u_i^l$ and naturally appear in the derivatives of $f_k$ with respect to the parameters: 
\begin{align}
\begin{aligned}
\frac{\partial f_{k}}{\partial W_{ij}^l} &= \delta_{k,i}^l h_j^{l-1}, \ \ \frac{\partial f_{k}}{\partial b_{i}^l} = \delta_{k,i}^l,
\\ \delta_{k,i}^{l} &= \phi'(u_i^l) \sum_{j} \delta_{k,j}^{l+1} W_{ji}^{l+1}.
\label{b_chain0}
\end{aligned}
\end{align}
To avoid complicating the notation, we will omit the index $k$ of the output unit, i.e., $\delta_{i}^{l} = \delta_{k,i}^{l}$. To evaluate the above feedforward and backward signals, we assume the following conditions. 

{\bf Random weights and biases:}
Suppose that the parameter set  is an ensemble generated by 
\begin{equation}
W_{ij}^l \overset{\text{i.i.d.}}{\sim}\mathcal{N}(0,\sigma_{w}^2/M_{l-1}), \ \ 
b_i^l \overset{\text{i.i.d.}}{\sim}\mathcal{N}(0,\sigma_{b}^2), \label{eq2}
\end{equation}
and thus is fixed, where $\mathcal{N}(0,\sigma^2)$ denotes a Gaussian distribution with zero mean and variance $\sigma^2$. Treating the case in which different layers have different variances is straightforward. Note that the variances of the weights are scaled in the order of $1/M$. In practice, the learning of DNNs usually starts from random initialization with this scaling \cite{glorot2010understanding,He_2015_ICCV}. 

{\bf Input samples:}
We assume that there are $N$ input samples $x(n) \in \mathbb{R}^{M_0}$ ($n=1,...,N$) generated identically and independently from the input distribution. We generate the samples by using a standard normal distribution, i.e.,
\begin{equation}
 x_j(n) \overset{\text{i.i.d.}}{\sim} \mathcal{N}(0,1). \label{eq7:0410} 
\end{equation}

{\bf Activation functions:}
Suppose the following two conditions: (i) {\it the activation function $\phi(x)$ has a polynomially bounded weak derivative}. (ii) {\it the network is non-centered}, which means a DNN with bias terms ($\sigma_b \neq 0$) or activation functions satisfying a non-zero Gaussian mean. The definition of the non-zero Gaussian mean is $\int Dz \phi(z) \neq 0$.
 The notation $Du = du \exp(-u^2/2) /\sqrt{2\pi}$ means integration over the standard Gaussian density.
 
Condition (i) is used to obtain recurrence relations of backward order parameters \cite{yang2019scaling}. Condition (ii) plays an essential role in our evaluation of the FIM  \cite{karakida2018universal,karakida2019normalization}. The two conditions are valid in various realistic settings, because conventional networks include bias terms, and widely used activation functions, such as the sigmoid function and (leaky-) ReLUs, have bounded weak derivatives and non-zero Gaussian means. Different layers may have different activation functions.

\begin{figure}[t]
\vskip 0.2in
\begin{center}
\centerline{\includegraphics[width=0.7\columnwidth]{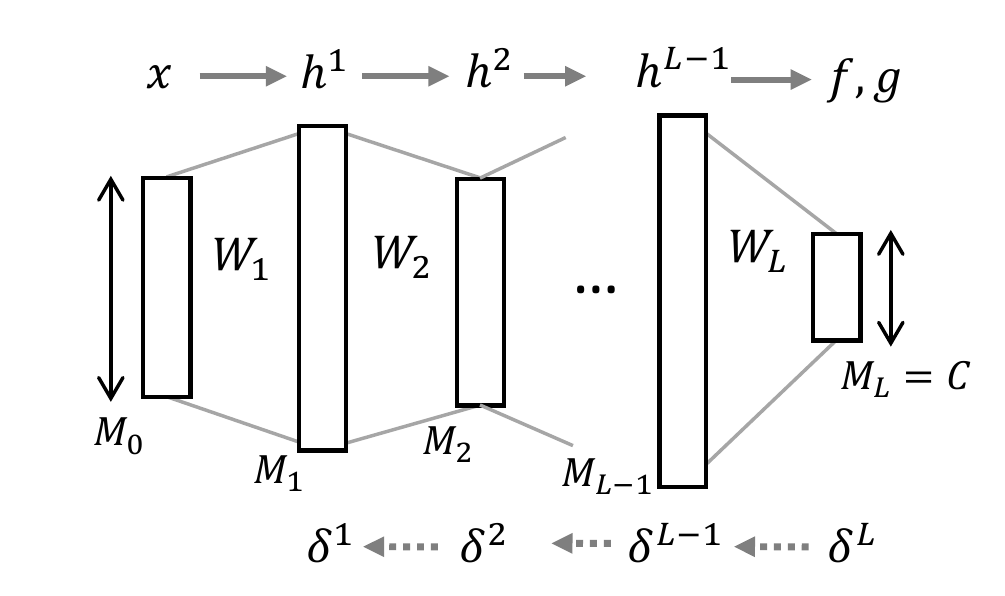}}
\caption{Deep neural networks (DNNs). The mathematical definitions are given in Section 2.1.}
\end{center}
\vskip -0.2in
\end{figure}

\subsection{Overview of metric tensors}
We will analyze two types of metric tensors (metric matrices) that determine the responses of network outputs, i.e., the response to a local change in parameters and the response to a local change in the input and hidden neurons. They are summarized in Fig. 2. One can systematically understand these tensors from the perspective of perturbations of variables.

We denote the set of network parameters as $\theta \in \mathbb{R}^P$. $P$ is the number of parameters. Next suppose we choose one network output unit $k$. If $f_k$ is perturbed by an infinitesimal change $d\theta$, its change is given by a quadratic form after performing a Taylor expansion, i.e., 
\begin{equation}
\mathrm{E} \left[ ||f(x;\theta + d\theta) - f(x;\theta)||^2 \right] \sim d\theta^{\top} F d\theta, \label{eq7:0323}
\end{equation}
\begin{equation}
 F := \sum_{k=1}^C \mathrm{E} \left[  \nabla_\theta f_k(x) \nabla_\theta f_k(x)^{\top} \right], \label{eq9:0816}
\end{equation}
where  $||\cdot ||$ denotes the Euclidean norm, $\mathrm{E}[\cdot]$ denotes the expectation over an input distribution, and $\nabla_\theta$ is the derivative with respect to $\theta$.
The matrix $F$ acts as {\it a metric tensor for the parameter space}. $F$'s eigenvalues determine the robustness of the network output $f$ against the perturbation. As will be explained in Section \ref{sec3_1}, this $F$
has a special meaning because it is the {\it Fisher information matrix (FIM)} corresponding to the MSE loss. 

When $N$ input samples $x(n)$ $(n=1,...,N)$ are available, we can replace the expectation $\mathrm{E}[\cdot]$ of the FIM with the empirical mean:
\begin{equation}
F= \sum_{k=1}^C \frac{1}{N} \sum_{n=1}^{N} \nabla_\theta f_k(n) \nabla_\theta f_k(n)^{\top}, \label{eqFIM}
\end{equation}
where we have abbreviated the network outputs as $f_k(n)=f_k(x(n);\theta)$ to avoid complicating the notation.  This is {\it an empirical FIM} in the sense that the average is computed over empirical input. 
We can express it in the matrix form shown in Fig. 2(a).  The Jacobian $ \nabla_\theta f$ is a $P \times CN$ matrix whose each column corresponds to $\nabla_\theta f_k(n)$ ($k=1,...,C$, $n=1,...,N$).  We investigate this type of empirical metric tensor for arbitrary $N$.
One can set $N$ as a constant value or make it increase depending on $M$.  
The empirical FIM (\ref{eqFIM}) converges to the expected FIM as $N \rightarrow \infty$. In addition, 
the FIM can be partitioned into $L^2$ layer-wise block matrices. We denote the $(l,l')$-th block as $F^{ll'}$ ($l,l'=1,...,L$).
We take a closer look at the eigenvalue statistics of diagonal blocks  in Section \ref{Sec_diag}. 

In this paper, we also investigate another FIM denoted by $F_{cross}$ which corresponds to 
classification tasks with cross-entropy loss.  
As shown in Fig. 2(a), 
one can represent $F_{cross}$ as a modification of $F$. A specific coefficient matrix $Q$ is inserted between the Jacobian $\nabla_\theta f$ and its transpose. $Q$ is defined by (\ref{eqFIM2}) and composed of nothing but softmax functions. Its mathematical definition is given in Section \ref{Sec_cross}. 
One more interesting quantity is a left-to-right reversed product of $\nabla_\theta f$ described in Fig. 2 (b). This matrix is known as the neural tangent kernel (NTK). The FIM and NTK share the same non-zero eigenvalues by definition, although we need to be careful in the change of parameterization used in the studies of NTK. The details are shown in Section 4.

In analogy with the FIM, one can introduce a metric tensor that measures the response to a change in the neural activities. Make a vector of all the activations in the input and hidden layers, i.e., $h:= \{h^0,h^1, ...,h^{L-1}\} \in \mathbb{R}^{M_h}$ with $M_h=\sum_{i=0}^{L-1} M_i$. Next, define an infinitesimal perturbation of $h$, i.e., $dh \in \mathbb{R}^{M_h}$, that is independent of $x$. Then, the response can be written as
\begin{equation}
\mathrm{E}[||f(h+dh;\theta)-f(h;\theta)||^2] \sim dh^{\top} A dh,
\end{equation}
 \begin{equation}
 A:= \mathrm{E} \left[\sum_{k=1}^C  \nabla_h f_k \nabla_h f_k^{\top} \right]. \label{eq0107:10}
 \end{equation}
We refer to $A$ as {\it the metric tensor for the input and feature spaces} because each $h^l$ acts as the input to the next layer and corresponds to the features realized in the network. 
We can also deal with a layer-wise diagonal block of $A$. Let us denote the $(l,l')$-th block by 
$ A^{ll'}$ ($l,l'=0,...,L-1$). 
In particular, the first diagonal block $A^{00}$ indicates the robustness of the network output against perturbation of the input: 
\begin{equation}
 \mathrm{E}[||f(x+dx;\theta)-f(x;\theta)||^2] \sim dx^{\top} A^{00} dx.
\end{equation}
Robustness against input noise has been investigated by similar (but different) quantities, such as sensitivity \cite{novak2018sensitivity}, and robustness against adversarial examples \cite{goodfellow2014explaining}. 

\begin{figure}[t]
\vskip 0.2in
\begin{center}
\centerline{\includegraphics[width=0.6\columnwidth]{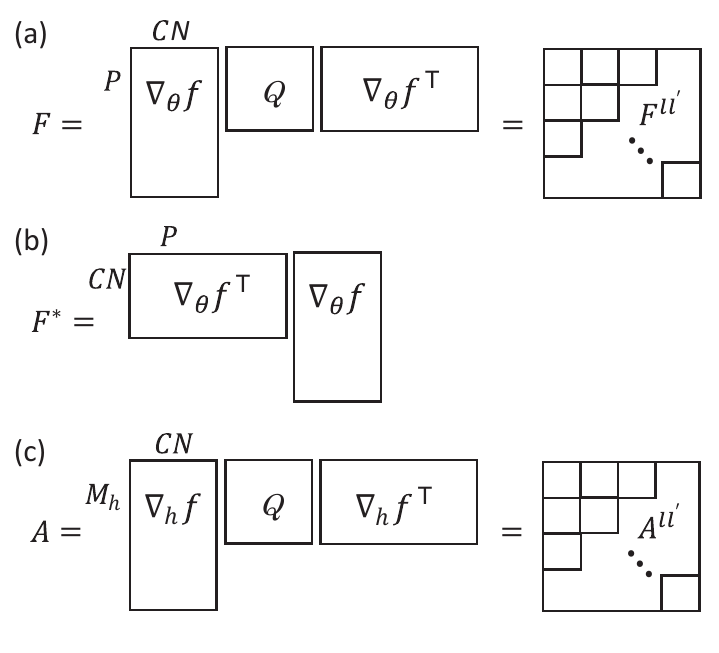}}
\caption{Matrix representations of metric tensors. (a) Metric for parameter space, also known as empirical Fisher information matrix (FIM). In particular, $Q=I$ corresponds to the FIM for MSE loss with linear output. (b) Dual of FIM. Under a specific parameter transformation, this is equivalent to the neural tangent kernel. (c) Metric for input and feature spaces. Note that the figures omit the scalar factors of the metrics.}
\end{center}
\vskip -0.2in
\end{figure}

\subsection{Order parameters for wide neural networks}
We use the following four types of {\it order parameter}, i.e., $(\hat{q}^l_{{1}},\hat{q}_{{2}}^l,\tilde{q}^l_{{1}},\tilde{q}_{{2}}^l)$, which were used in various studies on wide DNNs \cite{amari1974method,poole2016,schoenholz2016,yang2017,xiao2018dynamical,lee2017deep}. First, let us define the following variables for feedforward signal propagation; 
\begin{equation}
\hat{q}^l_{{1}} := \frac{1}{M_l} \sum_{i=1}^{M_l} h^{l}_i(n)^2, \ \ \hat{q}^l_{{2}} := \frac{1}{M_l} \sum_{i=1}^{M_l} h_i^{l} (n) h_{i}^{l}(m), \label{eq16:0410} 
\end{equation}
where $h_i^l(n)$ is the output of the $l$-th layer generated by the $n$-th input sample $x(n)$ ($n=1,...,N$). The variable $\hat{q}^l_{{1}}$ describes the total activity in the $l$-th layer, and the variable $\hat{q}^l_{{2}}$ describes the overlap between the activities for different input samples $x(n)$ and $x(m)$. These variables have been utilized to describe the depth to which signals can propagate from the perspective of order-to-chaos phase transitions \cite{poole2016}. In the large $M$ limit, these variables can be recursively computed by integration over Gaussian distributions \citep{poole2016,amari1974method}: 
\begin{align}
\begin{aligned}
 \hat{q}^{l+1}_{{1}} &= \int Du\phi ( \sqrt{q^{l+1}_{{1}}} u )^2, \ \ 
\hat{q}^{l+1}_{{2}} = I_{\phi}[q^{l+1}_{{1}},q_{{2}}^{l+1}], \\ 
q^{l+1}_{{1}} &:= \sigma_w^2 \hat{q}^l_{{1}} +\sigma_b^2, \ \ 
q_{{2}}^{l+1} :=\sigma_w^2 \hat{q}_{{2}}^l+\sigma_b^2, \label{eq_hatq}
\end{aligned}
\end{align}
for $l=0, ..., L-1$. Because the input samples generated by Eq. (\ref{eq7:0410}) yield $\hat{q}^0_{{1}} = 1$ and $\hat{q}_{{2}}^0 = 0$ for all $n$ and $m$, $\hat{q}_{{2}}^l$ in each layer takes the same value for all $n \neq m$; so does $\hat{q}_{{1}}^l$ for all $n$. A two-dimensional Gaussian integral is given by 
\begin{equation}
 I_{\phi}[a,b] := \int Dy Dx \phi (\sqrt{a}x)\phi (\sqrt{a}(cx +\sqrt{1-c^2}y )) 
\end{equation}
with $c=b/a$. One can represent this integral in a bit simpler form, i.e., $I_{\phi}[a,b] = \int Dy (\int Dx \phi (\sqrt{a-b}x +\sqrt{b}y ) )^2. $ 

Next, let us define the following variables for backpropagated signals:
\begin{equation}
 \tilde{q}^l_{{1}} := \sum_{i=1}^{M_l} \delta_i^l(n)^2, \ \ \tilde{q}^l_{{2}} := \sum_{i=1}^{M_l} \delta_i^{l} (n) \delta_{i}^{l}(m). \label{eq20:0410} 
\end{equation}
Above, we omitted $k$, the index of the output $f_k$, because the symmetry in the layer makes the above variables independent of $k$ in the large $M$ limit. 
Note that each $\delta_i^l$ is of $O(1/\sqrt{M})$ and their sums are of $O(1)$ in terms of the order notation $O(\cdot)$. The variable $\tilde{q}^l_{{1}}$ is the magnitude of the backward signals and $\tilde{q}^l_{{2}}$ is their overlap. Previous studies found that these order parameters in the large $M$ limit are easily computed using the following recurrence relations \cite{schoenholz2016,yang2019scaling}, 
\begin{equation} 
\tilde{q}^l_{{1}} = \sigma_w^2 \tilde{q}^{l+1}_{{1}} \int Du \phi' (\sqrt{q^{l}_{{1}}}u)^2, \ \
\tilde{q}_{{2}}^l = \sigma_w^2 \tilde{q}_{{2}}^{l+1} I_{\phi'}[q^l_{{1}},q_{{2}}^l], \label{eq_{{1}}ilqst} 
\end{equation}
for $l=0,...,L-1$. A linear network output (\ref{eq2:0323}) leads to the following initialization of the recurrences: $\tilde{q}^L_{{1}} = \tilde{q}_{{2}}^L  =1. $ The previous studies showed excellent agreement between these backward order parameters and experimental results \cite{schoenholz2016,yang2017,xiao2018dynamical}. Although those studies required the so-called {\it gradient independence assumption} to derive these recurrences, \citet{yang2019scaling} recently proved that such an assumption is unnecessary when condition (i) of the activation function is satisfied. 

These order parameters depend only on the type of activation function, depth, and the variance parameters $\sigma_w^2$ and $\sigma_b^2$. The recurrence relations for the order parameters require $L$ iterations of one- and two-dimensional numerical integrals. Moreover, we can obtain explicit forms of the recurrence relations for some of the activation functions \cite{karakida2018universal}.

\section{Eigenvalue statistics of FIMs}
\label{sec3}
This section shows the asymptotic eigenvalue statistics of the FIMs. When we have an $P \times P$ metric tensor whose eigenvalues are $\lambda_i$ ($i=1,..., P$), we compute the following quantities: 
\begin{equation}
 m_\lambda:= \frac{1}{P}\sum_{i=1}^{P} \lambda_i, \ \ s_\lambda:=\frac{1}{P} \sum_{i=1}^{P} \lambda_i^2, \ \ \lambda_{max}:=\max_i \lambda_i. \nonumber
\end{equation}
The obtained results are universal for any sample size $N$ which may depend on $M$. 

\subsection{FIM for regression tasks}
\label{sec3_1}
This subsection overviews the results obtained in the previous studies \cite{karakida2018universal,karakida2019normalization}. The metric tensor $F$ is equivalent to the Fisher information matrix (FIM) \cite{ama1998,pascanu2013,ollivier2015riemannian,park2000,martens2015optimizing}, originally defined by 
\begin{equation}
F:=\mathrm{E}\left[\nabla_\theta \log p(x,y;\theta) \nabla_\theta \log p(x,y;\theta)^{\top}\right]. \label{eq24:0323}
\end{equation}
The statistical model is given by $p(x,y;\theta)=p(y|x;\theta)q(x)$, where $p(y|x;\theta)$ is the conditional probability distribution of the DNN of output $y$ given input $x$, and $q(x)$ is an input distribution. The expectation $\mathrm{E}[\cdot]$ is taken over the input-output pairs $(x,y)$ of the joint distribution $p(x,y;\theta)$. This FIM appears in the Kullback-Leibler divergence between a statistical model and an infinitesimal change to it: $\mathrm{KL}[p(x,y;\theta):p(x,y;\theta+d\theta)] \sim d\theta^{\top} F d\theta.$ The parameter space $\theta$ forms a Riemannian manifold and the FIM acts as its Riemannian metric tensor \cite{amari2016information}. 

Basically, there are two types of FIM for supervised learning, depending on the definition of the statistical model. One type corresponds to the mean squared error (MSE) loss for regression tasks; the other corresponds to the cross-entropy loss for classification tasks. The latter is discussed in Section \ref{Sec_cross}. Let us consider the following statistical model for the regression:
 \begin{equation}
p(y|x;\theta) = \frac{1}{\sqrt{2\pi}} \exp \left( - \frac{1}{2}||y-f(x;\theta)||^2 \right).
 \end{equation}
 Substituting $p(y|x;\theta)$ into the original definition of FIM (\ref{eq24:0323}) and taking the integral over $y$, one can easily confirm that it is equivalent to the metric tensor (\ref{eq9:0816}) introduced by the perturbation.
Note that the  loss function $Loss(\theta)$ is given by the log-likelihood of this model, i.e., $Loss(\theta) = - \mathrm{E} [\ln p(y|x;\theta)]$, where we take the expectation over empirical input-output samples. 
This becomes the MSE loss.

The previous studies \cite{karakida2018universal,karakida2019normalization} uncovered the following eigenvalue statistics of the FIM (\ref{eqFIM}): 
\begin{theorem}[\cite{karakida2018universal},\cite{karakida2019normalization}]
{\it When $M$ is sufficiently large, the eigenvalue statistics of $F$ can be asymptotically evaluated as 
\begin{align}
m_\lambda &\sim \kappa_1 \frac{C}{M}, \ \ s_\lambda \sim \alpha \left(\frac{N-1}{N} \kappa_2^2 + \frac{\kappa_1^2}{N} \right)C, \nonumber
\end{align}
\begin{equation}
\lambda_{max}\sim \alpha \left( \frac{N-1}{N} \kappa_2 + \frac{\kappa_1 }{N}\right)M, \nonumber
\end{equation}
where $\alpha := \sum_{l=1}^{L-1} \alpha_l \alpha_{l-1}$, and positive constants $\kappa_1$ and $\kappa_2$ are obtained using order parameters, 
\begin{equation}
\kappa_1 := \sum_{l=1}^L\ \frac{\alpha_{l-1}}{\alpha} \tilde{q}^l_{{1}} \hat{q}^{l-1}_{{1}}, \ \ \kappa_2 := \sum_{l=1}^L \frac{\alpha_{l-1}}{\alpha} \tilde{q}^l_{{2}}\hat{q}^{l-1}_{{2}}. \nonumber
\end{equation}
}
\end{theorem}
The average of the eigenvalue spectrum asymptotically decreases in the order of $1/M$, while the variance takes a value of $O(1)$ and the largest eigenvalue takes a huge value of $O(M)$. It implies that most of the eigenvalues are asymptotically close to zero, while the number of large eigenvalues is limited. 
Thus, when the network is sufficiently wide, one can see that the shape of the spectrum asymptotically becomes pathologically distorted. 
This suggests that the parameter space of the DNNs is locally almost flat in most directions but highly distorted in a few specific directions.

In particular, regarding the large eigenvalues, we have
\begin{theorem}[Theorem 3.3 \cite{karakida2019normalization}]
{\it When $M$ is sufficiently large, 
the eigenspace corresponding to $\lambda_{max}$ is spanned by $C$ eigenvectors,
\begin{equation}
\mathrm{E}[\nabla_\theta f_k] \ \ (k=1,...,C). \nonumber
\end{equation}
When $N=\rho^{-1} M$ with a constant $\rho>0$, under the gradient independence assumption, 
 the second largest eigenvalue $\lambda_{max}'$ is bounded by \begin{equation}
   \rho \alpha (\kappa_1 - \kappa_2) +  c_1   \leq  \lambda_{max}'  \leq \sqrt{ (C\alpha^2 \rho (\kappa_1 - \kappa_2)^2 + c_2) M}, \label{eq19:0512}
\end{equation}
for non-negative constants $ c_1$ and $c_2$.
}
\end{theorem}
Since $\lambda_{max}'$ is of order $\sqrt{M}$ at most, the largest eigenvalues $\lambda_{max}$ act as outliers. It is noteworthy that batch normalization in the last layer can eliminate them. Such normalization includes mean subtraction, i.e., $\bar{f}_k := f_k-\mathrm{E}[f_k]$. The previous work \cite{karakida2019normalization} analyzed the corresponding FIM; 
\begin{align}
 \bar{F} &:= \sum_k \mathrm{E}[\nabla_\theta \bar{f}_k \nabla_\theta \bar{f}_k^\top ] \nonumber \\ 
 &= \sum_k (\mathrm{E}[\nabla_\theta {f}_k \nabla_\theta {f}_k^\top ] -  \mathrm{E}[\nabla_\theta {f}_k] \mathrm{E}[\nabla_\theta {f}_k ]^\top). \label{eq31:0626}
\end{align}
The subtraction (\ref{eq31:0626}) means eliminating the $C$ largest eigenvalues from $F$. 
Numerical experiments confirmed that the largest eigenvalue of $\bar{F}$ is of order 1. Note that when $N\propto M$, the sample size is sufficiently large but the network satisfies $P\gg N$ and keeps overparameterized. 

Figure 3 (a; left) shows a typical spectrum of the FIM. We computed the eigenvalues by using random Gaussian weights, biases, and inputs. We used deep Tanh networks with $L=3$, $M =200$, $C=10$, $\alpha_l=1$ and $(\sigma_w^2,\sigma_b^2)=(3,0.64)$. The sample size was $N=100$. The histograms were made from eigenvalues over $100$ different networks with different random seeds. The histogram had two populations. The red dashed histogram was made by eliminating the largest $C$ eigenvalues. It coincides with the smaller population. Thus, one can see that the larger population corresponds to the $C$ largest eigenvalues. The larger population in experiments can be distributed around $\lambda_{max}$ because $M$ is large but finite.

{{\bf Remark : Loss landscape and gradient methods.}}
The empirical FIM (\ref{eqFIM}) is equivalent to the Hessian of the loss, i.e., $\nabla_\theta^2 Loss(\theta)$, around the global minimum with zero training loss. 
\citet{karakida2019normalization} referred to the steep shape of the local loss landscape caused by $\lambda_{max}$ as {\it pathological sharpness}. The sharpness of the loss landscape is connected to an appropriate learning rate of gradient methods for convergence.  The previous work empirically confirmed that a learning rate $\eta$ satisfying
\begin{equation}
\eta<2/\lambda_{max} \label{eq24:0910}
\end{equation}
is necessary for the steepest gradient method to converge \cite{karakida2018universal}. 
In fact, this $2/\lambda_{max}$ acts as a boundary of neural tangent kernel regime \cite{lee2019wide,lewkowycz2020large}. 
Because $\lambda_{max}$ increases depending on the width and depth, we need to carefully choose an appropriately scaled learning rate to train the DNNs.

\subsection{Diagonal blocks of FIM}
\label{Sec_diag}
One can easily obtain insight into diagonal blocks, that is, $F^{ll}$, in the same way as Theorem 3.1.
Let us denote the set of parameters in the $l$-th layer by $\theta^l$. We have  $F^{ll}= \sum_k \mathrm{E}[\nabla_{\theta^l}f_k\nabla_{\theta^l}f_k^\top]$. 
 When $M$ is sufficiently large, the eigenvalue statistics of $F^{ll}$ are asymptotically evaluated as 
\begin{align}
m_\lambda^{l} &\sim \frac{\tilde{q}^l_{{1}} \hat{q}^{l-1}_{{1}}}{\alpha_l} \frac{C}{M}, \  s_\lambda^{l} \sim \frac{\alpha_{l-1}}{\alpha_l} \left(\frac{N-1}{N} (\tilde{q}^l_{{2}} \hat{q}^{l-1}_{{2}})^2 + \frac{(\tilde{q}^l_{{1}} \hat{q}^{l-1}_{{1}})^2}{N} \right)C, \nonumber
\end{align}
\begin{equation}
\lambda_{max}^{l} \sim \alpha_{l-1} \left( \frac{N-1}{N}\tilde{q}^l_{{2}} \hat{q}^{l-1}_{{2}} + \frac{\tilde{q}^l_{{1}} \hat{q}^{l-1}_{{1}} }{N}\right)M, \nonumber
\end{equation}
for $l=1,...,L$. 
The eigenspace corresponding to the largest eigenvalues is spanned by $C$ eigenvectors,
$\mathrm{E}[\nabla_ {\theta^l} f_k] \ \ (k=1,...,C).$
The derivation is shown in Appendix A.  The order of eigenvalue statistics is the same as the full-sized FIM. 
Figure  3(middle) empirically confirms that $F^{22}$ has a similar pathological spectrum to that of $F$. Its experimental setting was the same as in the case of $F$.

It is helpful to investigate the relation between $F$ and its diagonal blocks when one considers the diagonal block approximation of $F$. For example, use of a diagonal block approximation can decrease the computational cost of natural gradient algorithms \cite{martens2015optimizing,amari2018fisher}. When a matrix is composed only of diagonal blocks, its eigenvalues are given by those of each diagonal block.  $F$ approximated in this fashion has the same mean of the eigenvalues as the original $F$ and the largest eigenvalue $\max_l \lambda_{max}^l$, which is of $O(M)$. Thus, the diagonal block approximation also suffers from a pathological spectrum. Eigenvalues that are close to zero can make the inversion of the FIM in natural gradient unstable, whereas using a damping term seems to be an effective way of dealing with this instability \cite{martens2015optimizing}.

\begin{figure}
\vskip 0.2in
\begin{center}
\centerline{\includegraphics[width=1.00\columnwidth]{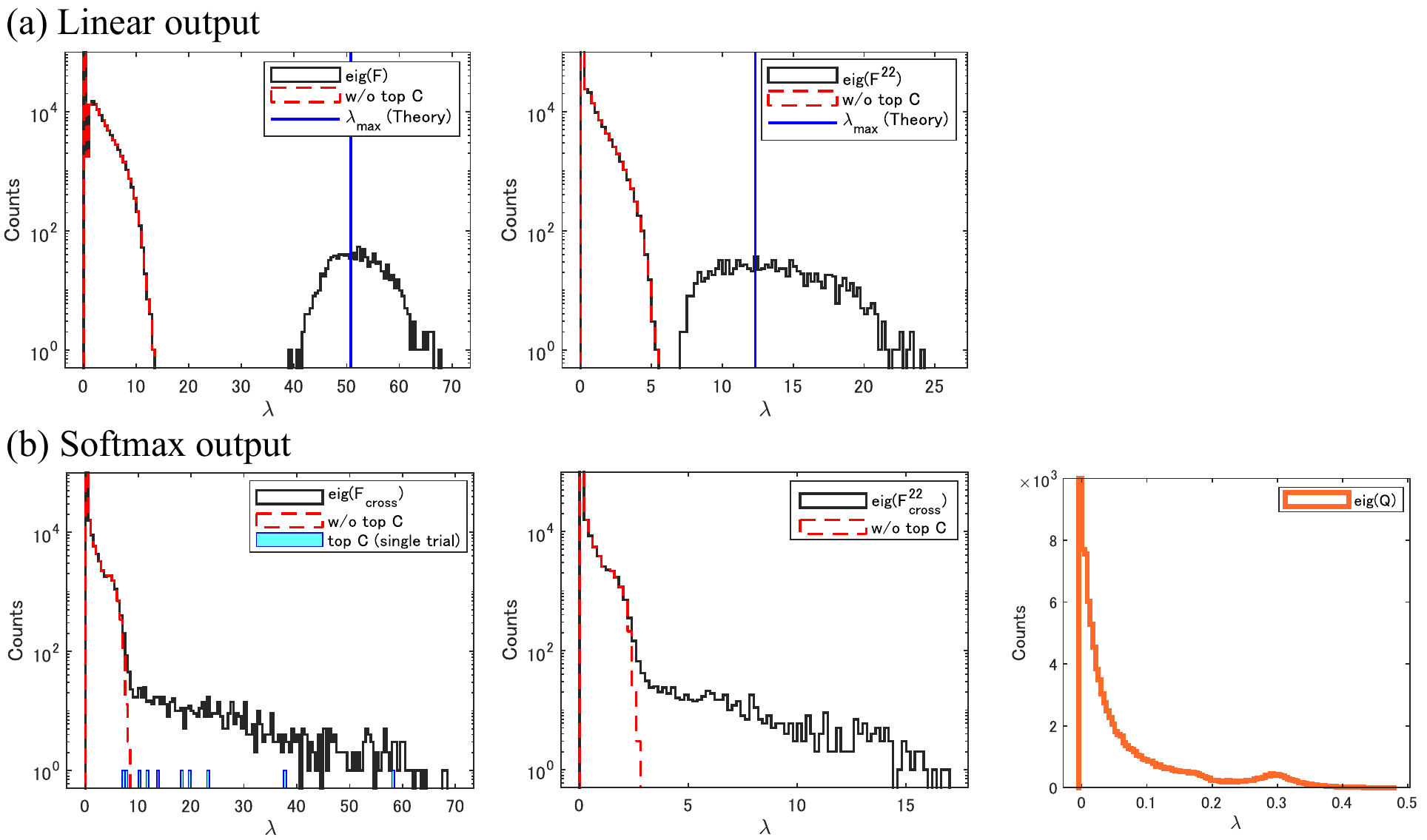}}
\caption{Eigenvalue spectra of FIMs in experiments with deep Tanh networks: (a; left) case of $F$, (a; right) case of diagonal block $F^{22}$. The vertical axis is the cumulative number of eigenvalues over $100$ different networks. The black histograms show the original spectra, while the red dashed ones show the spectra without the $C$ largest eigenvalues. The blue lines represent the theoretical values of the largest eigenvalues.  (b; left) case of $F_{cross}$, (b; middle) case of diagonal block $F_{cross}^{22}$, (b; right) case of $Q$. }
\end{center}
\vskip -0.2in
\end{figure}

\subsection{FIM for multi-label
classification tasks}
\label{Sec_cross}
The cross-entropy loss is typically used in multi-label classification tasks. It comes from the log-likelihood of the following statistical model: \begin{equation}
p(y|x;\theta) = \prod_{k=1}^C g_k (x)^{y_k},
 \end{equation}
where $g$ is the softmax output (\ref{eq25:0731}) and $y$ is a $C$-dimensional one-hot vector. The cross-entropy loss is given by $ -\mathrm{E}[ \sum_k y_k \log g_k]$. Substituting the statistical model into the definition of the FIM (\ref{eq24:0323}) and taking the summation over $y$, we find that the empirical FIM for the cross-entropy loss is given by
 \begin{equation}
F_{cross} = \frac{1}{N} \sum_{n}^{N} \sum_{k,k'}^C  \nabla_{\theta} f_k(n) Q_{{n}}(k,k') \nabla_{\theta}f_{k'}(n)^{\top}, \label{eqFIM2}
\end{equation}
 \begin{equation}
 Q_{{n}}(k,k') := \{g_k(n)\delta_{kk'} - g_k(n) g_{k'}(n)\}. \label{Qcross}
\end{equation}
The contribution of softmax output appears only in $Q_n$. $F_{cross}$ is linked to $F$ through the matrix representation shown in Fig. 2(a). One can view $F$ as a matrix representation with $Q=I$, that is, the identity matrix. In contrast, $F_{cross}$ corresponds to a block-diagonal $Q$ whose $n$-th block is given by the $C \times C$ matrix $Q_{n}$.
In a similar way to Eq. (\ref{eq7:0323}), we can 
see $F_{cross}$ as the metric tensor for the parameter space. Using the softmax output $g$, we have
\begin{equation}
4  \mathrm{E} \left[||\sqrt{g(x;\theta + d\theta)} - \sqrt{g(x;\theta)}||^2 \right] \sim d\theta^{\top} F_{cross} d\theta,
\end{equation}
where the square root is taken entry-wise.

We obtain the following result of $F_{cross}$'s eigenvalue statistics: 
\begin{theorem}
{\it When $M$ is sufficiently large, the eigenvalue statistics of $F_{cross}$ are asymptotically evaluated as 
\begin{equation}
m_\lambda \sim \beta_1 C \frac{\kappa_1}{M}, \ \
s_\lambda \sim \alpha \left( \beta_2 \kappa_2^2+\beta_3 \frac{\kappa_1^2}{N} \right), \nonumber
\end{equation}
\begin{equation}
 \beta_4 \alpha \left( \frac{N-1}{N} \kappa_2 + \frac{\kappa_1 }{N}\right)M \leq \lambda_{max} \leq \sqrt{\alpha s_{\lambda}} M, \nonumber
\end{equation}
where the constant coefficients are given by
\begin{align}
\beta_1&:= 1- \frac{\sum_n}{N}\sum_k^C g_k(n)^2, \nonumber \\
\beta_2 &:= \frac{\sum_{n \neq m}}{N^2} \left \{ \sum_k^C g_k(m) g_k(n) -2 \sum_k^C g_k(m)^2 g_k(n) \right. \nonumber  + \left. (\sum_k^Cg_k(m) g_k(n))^2 \right \},\nonumber \\
\beta_3 &:= \frac{\sum_n}{N} \left \{ \sum_k^C (1-2g_k(n))g_k(n)^2 + (\sum_k^Cg_k(n)^2)^2 \right \},\nonumber \\
\beta_4 &:= \max_{1 \leq k \leq C} \frac{\sum_n}{N} g_k(n) (1-g_k(n)). \nonumber
\end{align} 
}
\label{thm3_3}
\end{theorem}
The derivation is shown in Appendix B.1. 
We find that the eigenvalue spectrum shows the same width dependence as the FIM for regression tasks. Although the evaluation of $\lambda_{max}$ in Theorem 3.3 is based on inequalities, one can see that $\lambda_{max}$ linearly increases as the width $M$ or the depth $L$ increase. The softmax functions appear in the coefficients $\beta_k$. It should be noted that the values of $\beta_k$ generally depend on the index $k$ of each softmax output. This is because the values of the softmax functions depend on the specific configuration of $W^L$ and $b^L$.

If a relatively loose bound is acceptable, we can use the following simpler evaluation. 
Let us denote the eigenvalue statistics of $F$ shown in Theorem 3.1 by $(m_\lambda^{lin}, s_\lambda^{lin},\lambda_{max}^{lin})$. They corresponds to the contribution of linear output $f$ before putting it into the softmax function.
Taking into account the contribution of the softmax function, we have  
\begin{align}
m_\lambda &\leq  m_\lambda^{lin}, \ s_\lambda \leq  s_\lambda^{lin}, \
\lambda_{max} \leq  \lambda_{max}^{lin}.
\label{eq28:0914}
\end{align}
Note that $F_{cross}$'s eigenvalues satisfy $\lambda_i(F_{cross}) \leq \lambda_{max}(Q) \lambda_i(F)$ ($i=1,...,P$; $\lambda_1 \geq \lambda_2 \geq ... \geq \lambda_P$). The inequality (\ref{eq28:0914})  comes from $\lambda_{max}(Q) \leq 1$.  

\begin{figure}[t]
\begin{center}
\centerline{\includegraphics[width=0.9\columnwidth]{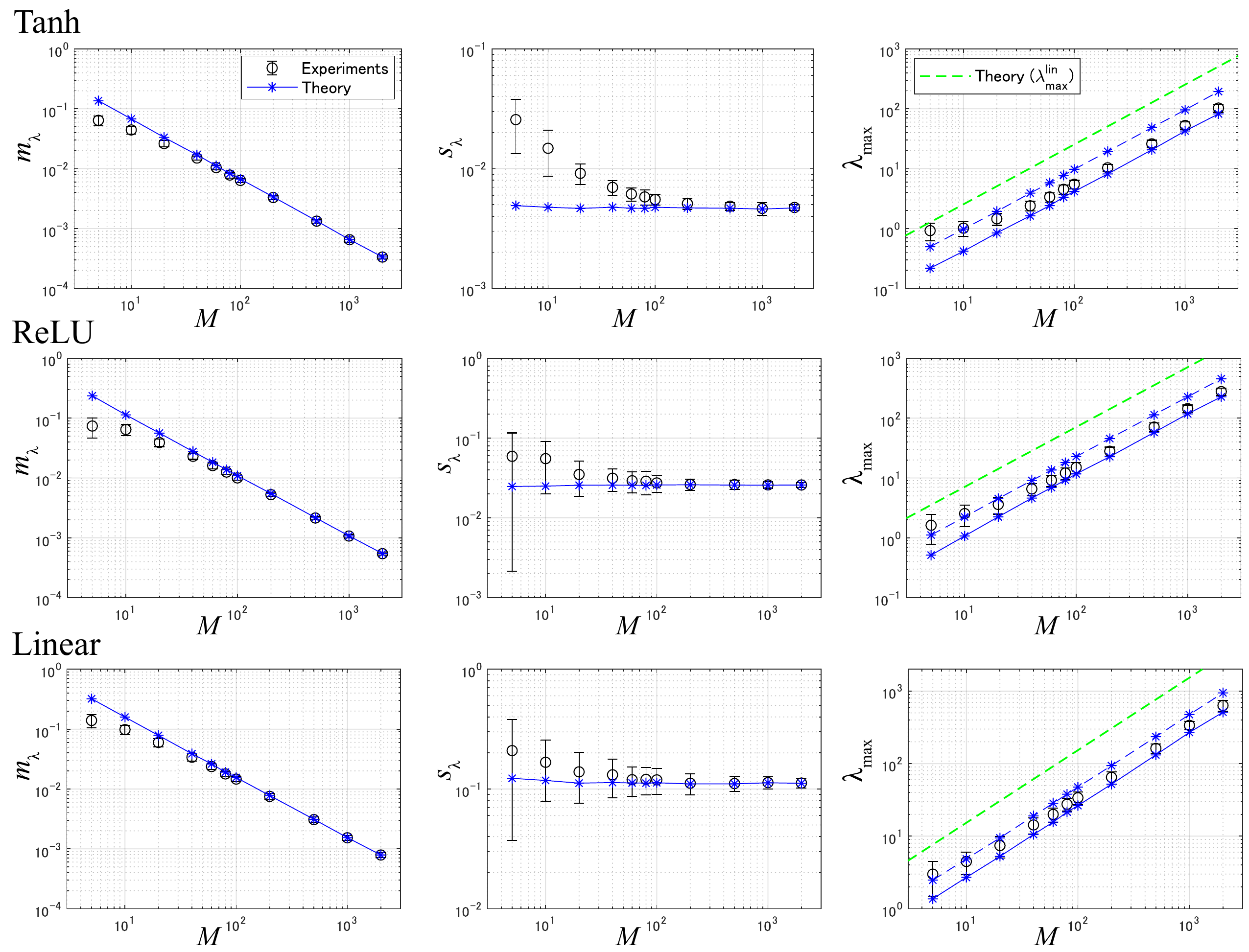}}
\caption{$F_{cross}$'s eigenvalue statistics: means (left), second moments (center), and maximum (right). Black points and error bars show means and standard deviations of the experimental results over 100 different networks with different random seeds. The blue lines represent the theoretical results obtained in the large $M$ limit. For $\lambda_{max}$, the dashed lines show the theoretical upper bound, while the solid ones show the lower bound.}
\end{center}
\end{figure}

Figure 4 shows that our theory predicts experimental results rather well for artificial data. We computed the eigenvalues of $F_{cross}$ with random Gaussian weights, biases, and inputs. We set $L=3$, $M =1000$, $C=10$, $\alpha_l=1$ and ($\sigma_w^2,\sigma_b^2$) = ($3,0.64$) in the tanh case, ($2,0.1$) in the ReLU case, and ($1,0.1$) in the linear case. The sample size was set to $N=100$. The predictions of Theorem 3.3 coincided with the experimental results for sufficiently large widths.

\subsection{$F_{cross}$'s large eigenvalues}

Exhaustive experiments on the cross-entropy loss have recently confirmed that there are $C$ dominant large eigenvalues (so-called outliers) \cite{papyan2019measurements}. Consistent with the results of this experimental study, we found that there are $C$ large eigenvalues: 
\begin{theorem}
{\it $F_{cross}$ has the first $C$ largest eigenvalues of $O(M)$. 
}
\end{theorem}
The theorem is proved in Appendix B.2. 
These $C$ large eigenvalues are reminiscent of the $C$ largest eigenvalues of $F$ shown in Theorem 3.1.

These $C$ largest eigenvalues can act as outliers. Note that because $\lambda_i(F_{cross}) \leq  \lambda_i(F)$, we have $\lambda_{C+1} \leq \lambda_{max}'$. $\lambda_{C+1}$ denotes 
the $(C+1)$-th largest eigenvalue of $F_{cross}$. 
Let us suppose that $N \propto M$ and the assumptions of Theorem 3.2 hold. In this case, $\lambda_{max}'$ is upper-bounded by (\ref{eq19:0512}) and then $\textbf{$\lambda_{C+1}$}$ is of $O(\sqrt{M})$ at most. This means that the first $C$ largest eigenvalues of $F_{cross}$ can become outliers.  
It would be interesting to extend the above results and theoretically quantify more precise values of these outliers. One promising direction will be to analyze a hierarchical structure of $F_{cross}$ empirically investigated by   \citet{papyan2019measurements}.   
It is also noteworthy that our outliers disappear under the mean subtraction of $f$ in the last layer mentioned in (\ref{eq31:0626}). This is because we have $\lambda_i(F_{cross}) \leq  \lambda_i(\bar{F})$ under the mean subtraction.

Figure 3(b; left) shows a typical spectrum of $F_{cross}$. We set 
$M=1000$ and other settings were the same as in the case of $F$. We found that compared to $F$, $F_{cross}$ had the largest $C(=10)$ eigenvalues which were widely spread from the bulk of the spectrum. 
Naively speaking, this is because the coefficient matrix $Q$ has distributed eigenvalues as is shown in Fig. 3(b; right). Compared to the FIM for regression, which corresponds to $Q=I$, 
the distributed $Q$'s eigenvalues can make $F_{cross}$'s eigenvalues disperse. Note that the black histogram is obtained as the summation over $100$ trials (different random initializations). The largest eigenvalues in a single trial are shown as blue boxes.  
 The diagonal block $F_{cross}$ also has the same characteristics of the spectrum as the original $F_{cross}$ in Fig. 3(b; middle).

Although this work mainly focuses on DNNs with random weights, it also gives some insight into the training of sufficiently wide neural networks. 
It is known that the whole training dynamics of gradient descent can be explained by NTK in sufficiently wide neural networks. See Section \ref{Sec4} for more details. In the case of the cross-entropy loss, its functional gradient is given by $ Q\Theta (y-g)$ \cite{lee2019wide}. $\Theta$ is the NTK at random initialization described in (\ref{eq63:0323}). $Q$ depends on the softmax function $g$ at each time step. One can numerically solve the training dynamics of $g$ and obtain the theoretical value of the training loss \cite{lee2019wide}. In Fig. 5(left), we confirmed that the theoretical line coincided well with the experimental results of gradient descent training. We set random initialization by $(\sigma_w^2,\sigma_b^2)=(2,0)$. We used artificial data with Gaussian inputs ($N=100$) and generated their labels by a teacher network whose architecture was the same  as the trained network. 
In addition, Figure 5(right) shows the largest eigenvalues during the training. 
As is expected from NTK theory, $\lambda_{max}^{lin}$ kept unchanged during the training.
In contrast, the $F_{cross}$'s largest eigenvalue dynamically changed because $Q$ depends on the softmax output $g$ which changes with the scale of $O(1)$. 
We calculated theoretical bounds of $\lambda_{max}$ (blue lines) by substituting $g$ at each time step into Theorem 3.3.     
Note that the bounds obtained in Theorem 3.3 are available even in the NTK regime because the coefficients $\beta_k$ admit any value of $g$ and are not limited to the random initialization. These theoretical bounds  explained well the experimental results of $\lambda_{max}$ during the training.  

Because the global minimum is given by $g=y$, all entries of $Q$ and $\lambda_{max}$ approach zero after a large enough number of steps. In the MSE case, we can explain the critical learning rate for convergence as is remarked in (\ref{eq24:0910}). In contrast, it is challenging to estimate such a critical learning rate in the cross-entropy case since $Q$ and $F_{cross}$ dynamically change.

\begin{figure}[t]
\begin{center}
\centerline{\includegraphics[width=0.9\columnwidth]{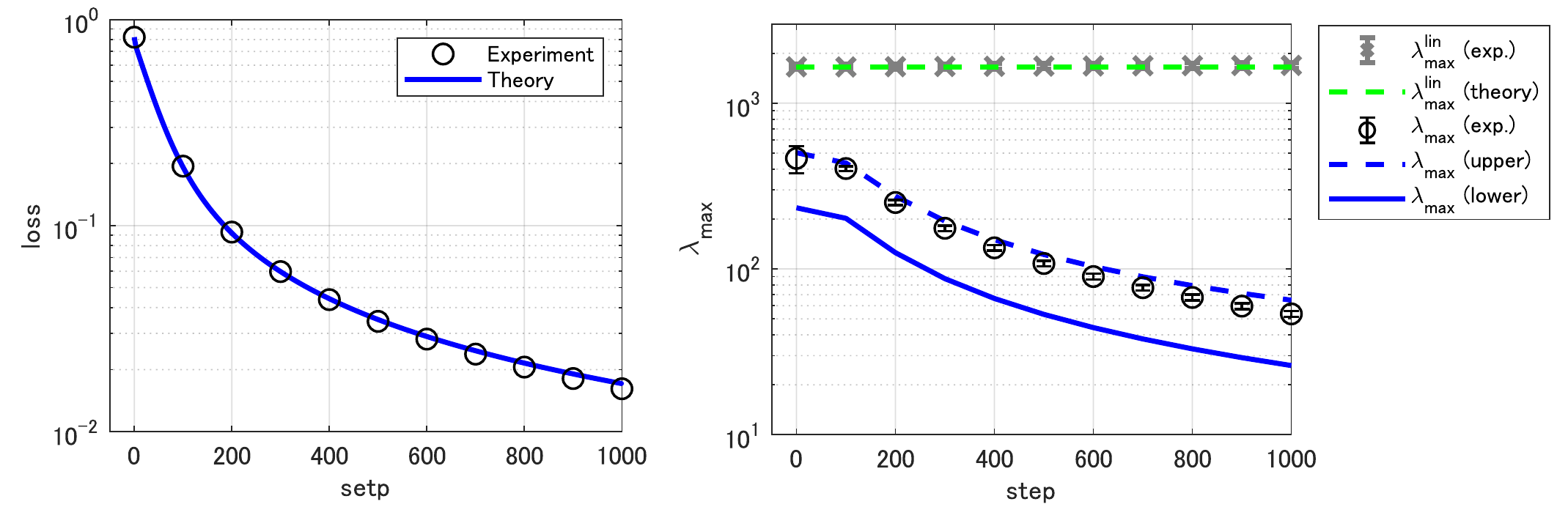}}
\caption{Training with the cross-entropy loss: we trained a 3-layered deep neural network ($L=3$, $M=2000$, $C=2$, $\alpha_l=1$) by gradient descent on artificial data. (left) training dynamics of the cross-entropy loss. (right) training dynamics of the largest eigenvalue. The largest eigenvalue of $F$ (grey crosses) keeps unchanged while that of $F_{cross}$ (black points)  dynamically changes and approaches zero.  We trained 5 different networks with different random seeds and showed its average and deviation. The colored lines represent the theoretical results obtained in the large $M$ limit.}
\end{center}
\end{figure}

\section{Connection to Neural Tangent Kernel}
\label{Sec4}
\subsection{Scale-dependent eigenvalue statistics}
The empirical FIM (\ref{eqFIM}) is essentially connected to a recently proposed Gram matrix, i.e., {\it the Neural Tangent Kernel (NTK)}.  \citet{jacot2018neural} defined the NTK by 
\begin{eqnarray}
\Theta := \nabla_\theta f^{\top} \nabla_\theta f. \label{eq63:0323}
\end{eqnarray}
Note that the Jacobian $ \nabla_\theta f$ is a $P \times CN$ matrix whose each column corresponds to $\nabla_\theta f_k(n)$ ($k=1,...,C$, $n=1,...,N$). 
Under certain conditions with sufficiently large $M$, the NTK at random initialization governs the whole training process in the function space by 
\begin{eqnarray}
\frac{df}{d t }&=& 
\frac{\eta}{N} \Theta (y-f), \label{dynamics_{{1}}tk}
\end{eqnarray} 
where the notation $t$ corresponds to the time step of the parameter update and $\eta$ represents the learning rate. Specifically, NTK's eigenvalues determine the speed of convergence of the training dynamics. Moreover, one can predict the network output on the test samples by using the Gaussian process with the NTK \cite{jacot2018neural,lee2019wide}.

The NTK and empirical FIM share essentially the same non-zero eigenvalues. It is easy to see that one can represent the empirical FIM (\ref{eqFIM}) by $F= \nabla_\theta f \nabla_\theta f^\top/N$. 
This means that the NTK (\ref{eq63:0323}) is the left-to-right reversal of $F$ up to the constant factor $1/N$.  \citet{karakida2018universal} introduced 
$F^* =  \nabla_\theta f^\top \nabla_\theta f/N$, which is essentially the same as the NTK,
and referred to $F^*$ as the dual of $F$. They 
used $F^*$ to derive Theorem 3.1.  

It should be noted that the studies of NTK typically suppose 
a special parameterization different from the usual setting. They 
consider DNNs with a parameter set
 $\theta = \{\omega_{ij}^l, \beta_i^l \}$ which determines weights and biases by 
\begin{equation}
W^l_{ij} = \frac{\sigma_w}{\sqrt{M_{l-1}}}\omega_{ij}^l, \ \ b^l_{i} = {\sigma_b}\beta_{i}^l, \ \ \omega_{ij}^l,\beta_{i}^l \overset{\text{i.i.d.}}{\sim} \mathcal{N}(0,1). 
\end{equation}
This NTK parameterization changes the scaling of Jacobian.
For instance, we have $\nabla_{\omega^l_{ij}} f =\frac{\sigma_w}{\sqrt{M_{l-1}}} \nabla_{W^l_{ij}} f$.
This makes the eigenvalue statistics slightly change from those of Theorem 3.1. 
 When $M$ is sufficiently large, the eigenvalue statistics of $\Theta$ under the NTK parameterization are asymptotically evaluated as 
\begin{align}
m_\lambda &\sim \alpha \kappa_1'C, \ \ s_\lambda \sim \alpha^2 \left( ({N-1}) \kappa_2'^2 + \kappa_1'^2 \right)C, \nonumber
\end{align}
\begin{equation}
\lambda_{max}\sim \alpha \left( (N-1) \kappa_2' + \kappa_1' \right). \nonumber
\end{equation}
The positive constants $\kappa_1$ and $\kappa_2$ are obtained using order parameters,
\begin{align}
\kappa_1' &:= \frac{1}{\alpha}\sum_{l=1}^L( \sigma_w^2 \tilde{q}^l_{{1}} \hat{q}^{l-1}_{{1}} + \sigma_b^2 \tilde{q}^l_{{1}}), \ \ 
\kappa_2' := \frac{1}{\alpha} \sum_{l=1}^L (\sigma_w^2 \tilde{q}^l_{{2}} \hat{q}^{l-1}_{{2}} + \sigma_b^2 \tilde{q}^l_{{2}}).\nonumber
\end{align}
The derivation is given in Appendix C. The NTK parameterization makes the eigenvalue statistics independent of the width scale. This is because the NTK parameterization maintains the scale of the weights but changes the scale of the gradients with respect to the weights. It also makes ($\kappa_1, \kappa_2$) shift to ($\kappa_1', \kappa_2'$). This shift occurs because the NTK parameterization makes the order of the weight gradients $\nabla_\omega f$ comparable to that of the bias gradients $\nabla_\beta f$. The second terms in $\kappa_1'$ and $\kappa_2'$ correspond to a non-negligible contribution from $\nabla_\beta f$. 

\comk{
While $m_\lambda$ is independent of the sample size $N$, $\lambda_{max}$ depends on it. This means that the NTK dynamics converge non-uniformly. Most of the eigenvalues are relatively small and the NTK dynamics converge more slowly in the corresponding eigenspace. In addition, a prediction made with the NTK requires the inverse of the NTK to be computed \cite{jacot2018neural,lee2019wide}. When the sample size is large, the condition number of the NTK, i.e., $\lambda_{max}/\lambda_{min}$, is also large and the computation with the inverse NTK is expected to be numerically inaccurate. 
}

\begin{figure}
\begin{center}
\centerline{\includegraphics[width=0.75\columnwidth]{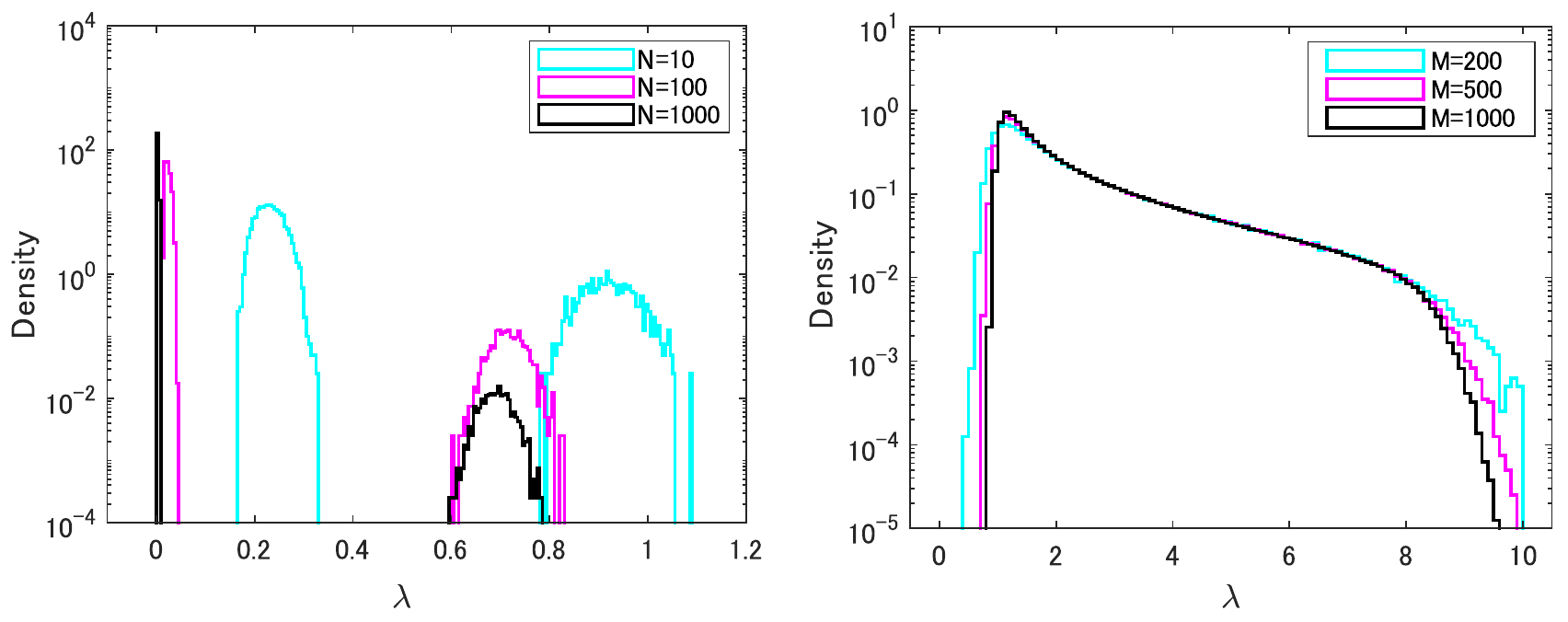}}
\caption{Spectra of the NTK ($\Theta$) in experiments with deep ReLU networks: (left) Spectra with $M=1000$ and various $N$. The eigenvalues are normalized by $1/N$ for comparison. (right) Spectra under mean subtraction in the last layer and the condition $N=M$. The vertical axes represent the probability density obtained from the cumulative number of eigenvalues over $400$ different networks.}
\end{center}
\end{figure}

\subsection{Scale-independent NTK}
A natural question is under what condition do NTK's eigenvalue statistics become independent of both the width and the sample size? As indicated in Eq. (\ref{eq31:0626}), the mean subtraction in the last layer with $N \propto M$ is a simple way to make the FIM's largest eigenvalue independent of the width. 
Similarly, one can expect that the mean subtraction makes the NTK's largest eigenvalue of $O(N)$ disappear and the eigenvalue spectrum take a range of $O(1)$ independent of the width and sample size. 

Figure 5 empirically confirms this speculation. We set $L=3$, $C=2$, $\alpha_l=1$ and used the Gaussian inputs and weights with $(\sigma_w^2,\sigma_b^2)=(2,0)$. As shown in Fig. 5 (left), NTK's eigenvalue spectrum becomes pathologically distorted as the sample size increases. To make an easier comparison of the spectra, the eigenvalues in this figure are normalized by $1/
N$. As the sample size increases, most of the eigenvalues concentrate close to zero while the
largest eigenvalues become outliers. In contrast, Figure  5 (right) shows that the mean subtraction keeps NTK's whole spectrum in the range of $O(1)$ under the condition $N \propto M$. The spectrum empirically converged to a fixed distribution in the large $M$ limit.

\section{Metric tensor for input and feature spaces}
\label{Sec5}

\begin{figure}
\begin{center}
\centerline{\includegraphics[width=0.70\columnwidth]{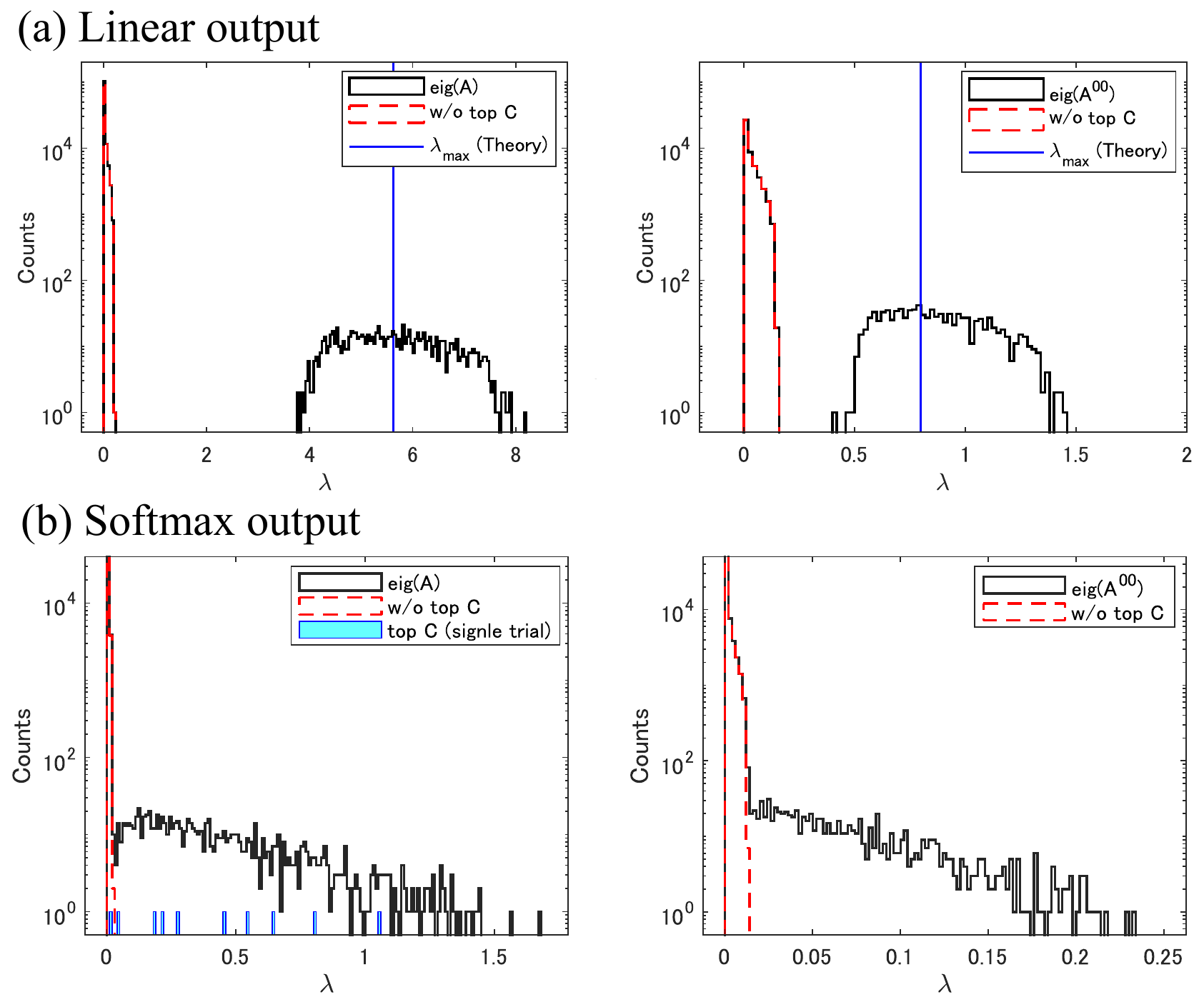}}
\caption{Spectra of metric tensor for input and feature spaces in experiments with deep Tanh networks: (left) the spectra of $A$, (right) the spectra of $A^{00}$. The vertical axis shows the cumulative number of eigenvalues over $100$ different networks. The black histograms show the original spectra, while the red dashed ones show the spectra without the $C$ largest eigenvalues. The blue lines represent the theoretical values of the largest eigenvalues. }
\end{center}
\end{figure}

The above framework for evaluating FIMs is also applicable to metric tensors for input and feature spaces, which are expressed in the matrix form in Fig. 2(c). Let us denote  $A_k:= \mathrm{E} \left[\nabla_h f_k \nabla_h f_k^{\top} \right]$. It is easy to see the eigenvalue statistics of $A$ from those of $A_k$. 
We can prove the following theorem:
\begin{theorem}
{\it When $M$ is sufficiently large, the eigenvalue statistics of $A_k$ are asymptotically evaluated as 
\begin{equation}
m_{\lambda} \sim \frac{\tilde{\kappa}_1}{M}, \ \ s_{\lambda} \sim \tilde{\alpha} \left ( \frac{N-1}{N} \tilde{\kappa}_2^2 + \frac{\tilde{\kappa}_1^2 }{N} \right ), \ \ \lambda_{max} \sim \tilde{\alpha} \left(\frac{N-1}{N} \tilde{\kappa}_2 + \frac{\tilde{\kappa}_1}{N}\right), \nonumber 
\end{equation}
where $\tilde{\alpha} := \sum_{l=0}^{L-1} \alpha_l$, and positive constants $\tilde{\kappa}_1$ and $\tilde{\kappa_2}$ are obtained from the order parameters, 
\begin{equation}
 \tilde{\kappa}_1 := \frac{\sigma_w^2}{\tilde{\alpha}} \sum_{l=1}^{L} \tilde{q}^l_{{1}}, \nonumber \ \ \tilde{\kappa}_2 := \frac{\sigma_w^2}{\tilde{\alpha}}\sum_{l=1}^{L} \tilde{q}^l_{{2}}. \nonumber
\end{equation}
The eigenvector of $A_k$ corresponding to $\lambda_{max}$ is $\mathrm{E}[\nabla_{h} f_k].$
}
\end{theorem}
The theorem is derived in Appendix D. Since $A$ is the summation of $A_k$ over $C$ output units, $m_\lambda$ and $s_\lambda$ of $A$ are $C$ times as large as those of $A_k$.  
The mean of the eigenvalues asymptotically decreases in the order of $O(1/M)$. 
Note that when $M_h \geq N$, $A_k$ has
trivial $M_h-N$ zero eigenvalues.
Even if we neglect these trivial zero eigenvalues, the mean becomes $ \tilde{\alpha} \tilde{\kappa}_1/N$ and decreases in the order of $O(1/N)$. 
In contrast, the largest eigenvalue is of $O(1)$ for any $M$ and $N$. 
Thus, the spectrum of $A_k$ is pathologically distorted in the sense that the mean is far from the edge beyond the order difference. The local geometry of $h$ is strongly distorted in the direction of $\mathrm{E}[\nabla_{h} f_k].$ 
Similarly, it is easy to derive the eigenvalue statistics of diagonal blocks $A^{ll}_k$. The details are shown in the Appendix.

Figure 7(a; left) shows typical spectra of $A$ and Figure 7(a; right) those of $A^{00}$. We used deep Tanh networks with $M=500$ and $N=1000$. The other experimental settings are the same as those in Fig. 3(a). The pathological spectra appear as the theory predicts. Similarly, we show spectra of softmax output in Fig. 7(b). 
The softmax output made the outliers widely spread in the same manner as in $F_{cross}$.

Let us remark on some related works in the literature of deep learning. First, \citet{pennington2018emergence} investigated similar but different matrices. Briefly speaking, they used random matrix theory and obtained the eigenvalue spectrum of $\nabla_{u^l} f$ with $N=1$, $M_l=C=M$. They found that the isometry of the spectrum is helpful to solve the vanishing gradient problem. Second, DNNs are known to be vulnerable to a specific noise perturbation, i.e., the adversarial example \cite{goodfellow2014explaining}. One can speculate that the eigenvector corresponding to $\lambda_{max}$ may be related to adversarial attacks, although such a conclusion will require careful considerations.

\section{Discussion}
We evaluated the asymptotic eigenvalue statistics of the FIM and its variants in sufficiently wide DNNs. They have pathological spectra in the conventional setting of random initialization and activation functions. This suggests that we need to be careful about the eigenvalue statistics and their influence on the learning when we use large-scale deep networks in naive settings.

The current work focused on fully-connected neural networks and it will be interesting to explore the spectra of other architectures such as ResNets and CNN. 
It will be also fundamental to explore the eigenvalue statistics that the current study cannot capture. While our study captured some of the basic eigenvalue statistics, it remains to derive the whole spectrum analytically. In particular, after the normalization excludes large outliers, the bulk of the spectrum becomes dominant. 
Random matrix theory enables us to analyze the FIM's eigenvalue spectrum in a shallow and centered network without bias terms \cite{pennington2018spectrum}. Extending the random matrix theory to more general cases including deep neural networks seems to be a prerequisite for further progress. 
Furthermore, we assumed a finite number of network output units. In order to deal with multi-label classifications with high dimensionality, it would be helpful to investigate eigenvalue statistics in the wide limit of both hidden and network output layers. Finally,
although we focused on the finite depth and regarded order parameters as constants,  
they can exponentially explode on extremely deep networks in the chaotic regime \cite{schoenholz2016,yang2019bn}. The NTK in such a regime has been investigated in \cite{jacot2019}.

It would also be interesting to explore further connections between the eigenvalue statistics and learning. Recent studies have yielded insights into the connection between the generalization performance of DNNs and the eigenvalues statistics of certain Gram matrices including FIM and NTK \cite{suzuki2018fast,Sun2019,yang2019fine}. We expect that the theoretical foundation of the metric tensors given in this paper will lead to a more sophisticated understanding and development of deep learning in the future.

\part*{Appendices}

\setcounter{section}{0}
\renewcommand{\thesection}{\Alph{section}}
\renewcommand{\theequation}{A.\arabic{equation} }
\setcounter{equation}{0}

\section{Eigenvalue statistics of $F$}
\subsection{Overviewing the derivation of Theorem 3.1}
The FIM is composed of feedforward and backpropagated signals. For example, its
 weight part is given by $(F^{ll'}_k)_{(ij)(i'j')}:= \mathrm{E} \left[ \nabla_{W^l_{ij}} f_k \nabla_{W^{l'}_{i'j'}} f_k \right]$. One can represent it in matrix form:
\begin{equation}
 F_k^{ll'} =\mathrm{E} [ \delta^l_k (\delta^{l'}_k)^{\top} \otimes h^{l-1} (h^{l'-1})^{\top} ], \label{eq0407:9}
\end{equation}
where $ \otimes$ represents the Kronecker product. The variables $h$ and $\delta$ are functions of $x$, and the expectation is taken over $x$.
This expression is not easy to treat in the analysis, and we introduce a dual expression of the FIM, which is essentially the same as NTK, as follows.

First, we briefly overview the derivation of Theorem 3.1 shown in  \cite{karakida2018universal,karakida2019normalization}. The essential point is that a Gram matrix has the same non-zero eigenvalues as its dual. One can represent the empirical FIM (\ref{eqFIM}) as 
\begin{align}
\begin{aligned}
F&=RR^{\top}, \\ 
R &:= 
\frac{1}{\sqrt{N}}[\nabla_\theta f_1 \ \ \nabla_\theta f_2 \ \ \cdots \ \ \nabla_\theta f_C]. \label{eq60:0111}
\end{aligned}
\end{align}
Its columns are the gradients on each input, i.e., $\nabla_{\theta} f_k(n)$ $(n=1,...,N)$. Let us refer to a $CN \times CN$ matrix $F^* :=R^{\top}R$ as the dual of FIM. Matrices $F$ and $F^*$ have the same non-zero eigenvalues by definition. This $F^*$ can be partitioned into $N \times N$ block matrices. The $(k,k')$-th block is given by 
\begin{equation}
 F^{*}(k,k') = \nabla_\theta f_k^{\top}\nabla_\theta f_{k'}/N,
\end{equation}
for $k,k'=1,...,C$. In the large $M$ limit, the previous study \cite{karakida2018universal} showed that $F^*$ asymptotically satisfies
\begin{equation}
 F^{*}(k,k') = \alpha \frac{M}{N} K \delta_{kk'}+ \frac{1}{N}o(M), \label{eq69:0629}
\end{equation}
where $\delta_{k,k'}$ is the Kronecker delta. As is summarized in Lemma A.1 in \cite{karakida2019normalization}, the second term of Eq. (\ref{eq69:0629}) is negligible in the large $M$ limit. In particular, it is reduced to $O(\sqrt{M})/N$ under certain condition. 
The matrix $K$ has entries given by
\begin{equation}
K_{nm}= \kappa_1 \ \ (n=m), \ \ \kappa_2 \ \ (n \neq m). 
\end{equation}
Using this $K$, the previous studies derived the basic eigenvalues statistics \cite{karakida2018universal} and eigenvectors corresponding to $\lambda_{max}$ \cite{karakida2019normalization}.
The matrix $K$ has the largest eigenvalue $( (N-1)\kappa_2 +\kappa_1 )$ and its eigenvectors $\nu_k \in \mathbb{R}^{CN}$ ($k=1,...,C$) whose entries are given by
\begin{align}
\begin{aligned}
 (\nu_k)_i := &\frac{1}{\sqrt{N}} \ \ ( (k-1)N+1 \leq i \leq kN), \ \\ 
 &\ \ 0 \ \ (\mathrm{otherwise}). \label{eq44:0323}
\end{aligned}
\end{align}
Note that $\kappa_1$ is positive by definition and $\kappa_2$ is positive under the condition (ii) of activation functions. 
The other eigenvalues of $K$ are given by $\kappa_1-\kappa_2$.

 We can obtain $m_\lambda$ from  $\mathrm{Trace}(F^*(k,k))C/P$,  $s_\lambda$ from  $ ||F^*(k,k)||_F^2C/P$ where $||\cdot||_F$ is the Frobenius norm, and $\lambda_{max}$ from $\nu_k^\top F^*(k,k) \nu_k$. The eigenvector of $F$ corresponding to $\lambda_{max}$ is asymptotically given by 
$\mathrm{E}[\nabla f_k]=R\nu_k$.
When $N$ is of $O(1)$,
it is obvious that $K$'s eigenvalues determine $F^*$'s eigenvalues in the large $M$ limit. 
Even if $N$ increases depending on $M$, our eigenvalue statistics hold in the large $M$ and $N$ limits. That is, we have asymptotically $m_\lambda \sim \kappa_1 C/M$, $s_\lambda \sim  \alpha \kappa_2^2 C$,  and $\lambda_{max} \sim \alpha \kappa_2 M$.
As one can see here, the condition of  $\kappa_2>0$ is crucial for our eigenvalue statistics. Non-centered networks guarantee $\hat{q}_{{2}}^l>0$ and $\tilde{q}_{{2}}^l>0$ which leads to $\kappa_2>0$.
In centered networks, $\kappa_2$ can become zero and we need to carefully evaluate the second term of Eq.~(\ref{eq69:0629}).

\subsection{Diagonal blocks}
We can immediately derive eigenvalue statistics of diagonal blocks in the same way as Theorem 3.1. We can represent the diagonal blocks as $F^{ll}:= R^lR^{l\top}$ with
\begin{eqnarray}
R^l &:=& \frac{1}{\sqrt{N}} 
[\nabla_{\theta^l} f_1 \ \ \nabla_{\theta^l} f_2 \ \ \cdots \ \ \nabla_{\theta^l} f_C],
\end{eqnarray}
and the dual of this Gram matrix as
\begin{eqnarray}
F^{ll*} &:=& R^{l\top}R^l,
\end{eqnarray}
where the parameter set $\theta^l$ means all parameters in the $l$-th layer. The $CN \times CN$ matrix $F^{ll*}$ can be partitioned into $N \times N$ block matrices whose $(k,k')$-th block is given by
\begin{equation}
 F^{ll*}(k,k') = \nabla_{\theta^l} f_k^{\top}\nabla_{\theta^l} f_{k'}/N,
\end{equation}
for $k,k'=1,...,C$. As one can see from the additivity of $\sum_{l=1}^L F^{ll*}(k,k')=F^{*}(k,k')$, the following evaluation is part of Eq. (\ref{eq69:0629}):
\begin{equation}
 F^{ll*}(k,k') = \alpha_{l-1} \frac{M}{N} K^{l} \delta_{kk'}+ \frac{1}{N}o({M}), 
\end{equation}
where
\begin{equation}
K^{l}_{nm}:= \tilde{q}^l_1\hat{q}^{l-1}_1 \ \ (n=m) ,\ \ \tilde{q}^l_{{2}}\hat{q}^{l-1}_{{2}} \ \ (n \neq m).
\end{equation}
Thus, we have $m_\lambda \sim \mathrm{Trace}(\alpha_{l-1} \frac{M}{N}K^l)C/P_l$ and $s_\lambda \sim  ||\alpha_{l-1} \frac{M}{N}K^l||_F^2C/P_l$, where the dimension of $\theta^l$ is given by  $P_l=\alpha_{l}\alpha_{l-1}M^2$. 
We set $\alpha_L = C/M$ in the last layer. 
The matrices $K$ and $K^l$ have the same eigenvectors corresponding to the largest eigenvalues, i.e., $\nu_k$. The largest eigenvalue is given by $\lambda_{max} \sim \alpha_{l-1} \frac{M}{N} \nu_k^\top K^l \nu_k$. The eigenvectors of $F^{ll}$ corresponding to $\lambda_{max}$ are $R^l \nu_k=\mathrm{E}[\nabla_{\theta^l} f_k]$. 

\section{Eigenvalue statistics of $F_{cross}$}
\subsection{Derivation of Theorem 3.3}
\renewcommand{\theequation}{B.\arabic{equation} }
\setcounter{equation}{0}
$F_{cross}$ is expressed by 
\begin{equation}
 F_{cross} := R Q R^{\top},
\end{equation}
where $Q$ is a $CN \times CN$ matrix. One can rearrange the columns and rows of $Q$ and partition it into $N \times N$ block matrices $Q (k,k')$ whose entries are given by 
\begin{equation}
 Q (k,k')_{nm} = \{g_k(n)\delta_{kk'} - g_k(n) g_{k'}(n)\} \delta_{nm},
\end{equation}
for $k,k'=1,...,C$. Each block is a diagonal matrix. 
Note that the non-zero eigenvalues of $RQR^{\top}$ are equivalent to those of $QR^{\top}R$. Since we have $F^*=R^{\top}R$, we should investigate the eigenvalues of the following matrix: 
 \begin{equation}
 F_{cross}^* := QF^*. \label{eqB3}
\end{equation}

The mean of the eigenvalues is given by
\begin{align}
m_{\lambda} &= \mathrm{Trace} (F_{cross}^*)/P \nonumber \\
&= \sum_{i,k} \mathrm{Trace}(Q(k,i) F^*(i,k))/P\nonumber \\ 
&\sim \sum_{k} \mathrm{Trace}(Q(k,k) F^*(k,k))/P\nonumber \\
&\sim C (1-\beta_1) \kappa_1/M.
\end{align} 
The third line holds asymptotically, since the order of $F(k,k)$ in Eq.~(\ref{eq69:0629}) is higher than that of $F^*(k,k')$ ($k\neq k'$). The fourth line comes from $\sum_k g_k(n)=1$. 

The second moment is evaluated as 
\begin{align}
s_{\lambda} &= \mathrm{Trace}(F_{cross}^{*2})/P \nonumber \\
&= \sum_{k} \mathrm{Trace} (\sum_{a,b,c}Q(k,a)F^*(a,b)Q(b,c)F^*(c,k) )/P \nonumber \\
&\sim \sum_{k,k'} \mathrm{Trace} (Q(k,k')F^*(k',k')Q(k',k)F^*(k,k) )/P.
\end{align}
Substituting $K$ into $F^*(k,k)$ gives
\begin{align}
s_{\lambda}&\sim \frac{\alpha}{N^2} \sum_{k,k'} \sum_{n} \left ( Q_n(k,k') ( \kappa_2^2 \sum_{m \setminus \{n\}}Q_m(k,k')  +\kappa_1^2Q_n(k,k') ) \right) \nonumber \\
&= \alpha \left( \beta_2 \kappa_2^2+\beta_3 \frac{\kappa_1^2}{N} \right),
\end{align} 
where $\sum_{m \setminus \{n\}}$ means a summation over $m$ excluding the $n$-th sample.

Finally, we derive the largest eigenvalue. Let us denote the eigenvectors of $F$ as 
\begin{equation}
 v_k := \frac{\mathrm{E}[\nabla f_k]}{||\mathrm{E}[\nabla f_k]||}. 
\end{equation}
It is easy to confirm that we have asymptotically $||\mathrm{E}[\nabla f_k]||^2 \sim  \lambda_{max}(F)$ \cite{karakida2019normalization}, where the largest eigenvalue of $F$ is denoted as $\lambda_{max}(F) = \alpha(\frac{N-1}{N}\kappa_2 + \frac{\kappa_1}{N} )M $. By definition, $F_{cross}$'s largest eigenvalue satisfies $\lambda_{max}\geq x^{\top} F_{cross}x$ for any unit vector $x$. By taking $x=v_k$, we obtain 
\begin{align}
 \lambda_{max} &\geq \lambda_{max}(F)^{-1} \cdot (R\nu_k)^{\top} F_{cross} (R\nu_k) \nonumber \\ 
 &= \lambda_{max}(F)^{-1} \cdot (F^*\nu_k)^{\top} Q (F^* \nu_k). 
\end{align}
Because we have asymptotically $F^*\nu_k = \lambda_{max}(F) \nu_k$, the lower bound is given by 
\begin{align}
 \lambda_{max}(F) &\geq \lambda_{max}(F) \cdot (\nu_k^{\top} Q\nu_k) \nonumber \\
 &= \lambda_{max}(F) \cdot \frac{1}{N} \sum_n g_k(n)(1-g_k(n)).
\end{align}
Taking the index $k$ that maximizes the right-hand side, we obtain the lower bound of $\lambda_{max}$. 
The upper bound of $\lambda_{max}$ immediately comes from a simple inequality for non-negative variables, i.e., $\lambda_{max} \leq \sqrt{\sum_i \lambda^2_i}= \sqrt{Ps_{\lambda}}$. Thus, we obtain Theorem 3.3. 

Note that we immediately have $\lambda_i(F^*_{cross}) \leq \lambda_{max}(Q)\lambda_i(F^*)$ from (\ref{eqB3}).
This means that $F_{cross}$'s eigenvalues satisfy $\lambda_i(F_{cross}) \leq \lambda_{max}(Q) \lambda_i(F)$ ($i=1,...,P$).
In addition, we have  $\lambda_{max}(Q) \leq 1$  because $\lambda_{max}(Q) \leq \max_n \lambda_{max}(Q_n)$ and $ \lambda_{max}(Q_n) \leq \max_k g_k(n) \leq 1$.  Therefore, $\lambda_i(F_{cross}) \leq \lambda_i(F)$ holds and we obtain inequalities (\ref{eq28:0914}).

\subsection{Derivation of Theorem 3.4}
Define $u_i$ to be the eigenvector of $F_{cross}$ corresponding to the eigenvalue $\lambda_i$ ($\lambda_1 \geq \cdots \geq \lambda_{i} \geq \cdots \geq \lambda_{P}$). Moreover, let us denote the linear subspace spanned by $\{u_1,...,u_k\}$ as $U_k$ and the orthogonal complement of $U_k$ as $U_k^\bot$. When $k=0$, we have $U_0^\bot= \mathbb{R}^P$. The dimension of $U_k^\bot$ is $P-k$, and we denote it as $\mathrm{dim}(U_k^\bot)=P-k$. Thus, we have 
\begin{align}
 \lambda_{r} &= \max_{||x||=1; x \in U_{r-1}^\bot} x^\top F_{cross} x, 
\end{align}
for $r=1,...,P$. Define $V_k$ to be a linear subspace spanned by $k$ eigenvectors of $F$ corresponding to $\lambda_{max}(F)$, i.e., $\{v_{i_1}, ..., v_{i_k}\}$. The indices $\{i_1, ...,i_{k}\}$ are chosen from $\{1,...,C\}$ without duplication. 

It is trivial to show from the dimensionality of the linear space that the intersection $S_r:= \{U_{r-1}^\bot \cap V_C\}$ is a linear subspace satisfying $C-r+1 \leq \mathrm{dim}(S_r) \leq C$ when $ 1\leq r \leq C$. Let us take a unit vector $x$ in $S_r$ as $x= \sum_{s=1}^{r^*} a_s v_{i_s}$, where we have defined $r^*:=\mathrm{dim}(S_r)$ and the coefficients $a_s$ satisfy $\sum_{s=1}^{r^*} a_s^2=1$. In the large $M$ limit, we asymptotically have 
\begin{align}
 \lambda_{r} &\geq \max_{||x||=1; x \in S_r} x^\top F_{cross} x \nonumber \\ 
 &= \max_{(a_1,...,a_{r^*});\sum a_s^2=1} \sum_{s,s'} a_{s} a_{s'} (\nu_{i_s}^\top Q \nu_{i_{s'}}) \cdot \lambda_{max}(F) \nonumber \\ 
 & \geq \nu_{i_1}^\top Q \nu_{i_1} \cdot \lambda_{max}(F) \nonumber \\ 
 &= \frac{1}{N} \sum_{n} g_{i_1}(n)(1-g_{i_1}(n))\cdot \lambda_{max}(F),
\end{align}
where $\lambda_{max}(F)$ is of $O(M)$ from Theorem 3.1. This holds for all of $r=1,...,C$ and we can say that there exist  $C$ large eigenvalues of $O(M)$. 

\section{Derivation of NTK's eigenvalue statistics}
\renewcommand{\theequation}{C.\arabic{equation} }
\setcounter{equation}{0}
The NTK is defined as $\Theta = N F^*$ under the NTK parameterization. 
In the same way as Eq. (\ref{eq69:0629}), the $(k,k')$-th block of the NTK is asymptotically given by 
\begin{equation}
 \Theta(k,k') = \alpha K' \delta_{kk'}+ o(1), \label{eqD1:0930}
\end{equation}
for $k,k'=1,...,C$. In contrast to Eq. (\ref{eq69:0629}), the NTK parameterization makes $F^*$ multiplied by $1/M$. The negligible term of $o(1)$ is reduced to $O(1/\sqrt{M})$ under the condition summarized in  \cite{karakida2019normalization}. 
The entries of $K'$ are given by 
\begin{equation}
K'_{nm}:= \kappa_1' \ \ (n=m), \ \ \kappa_2' \ \ (n \neq m),
\end{equation}
where $\kappa'_1$ is composed of two parts. The first part is $ ||\nabla_\omega f_k(n)||^2 = \sigma_w^2 \sum_{l,i,j} \delta_i^{l}(n)^2 h_j^{l-1}(n)^2/M_{l-1} \sim \sigma_w^2 \sum_{l=1}^L \tilde{q}^l_{{1}} \hat{q}^{l-1}_{{1}}$. The second term is $||\nabla_\beta f_k(n)||^2 = \sigma_b^2 \sum_{l,i} \delta_i^{l}(n)^2 \sim \sigma_b^2 \sum_{l=1}^L \tilde{q}^l_{{1}}$. Despite that the number of weights is much larger than the number of biases, the NTK parameterization makes a contribution of $\nabla_\omega f_k$ comparable to that of $\nabla_\beta f_k$. This is in contrast to the evaluation of $F^*$ in Theorem 3.1, where the contribution of $\nabla_b f_k$ is negligible \cite{karakida2018universal}. We can evaluate $\kappa'_2$ in the same way. 

In the same way as with the FIM, the trace of $K'$ leads to $m_\lambda$, the Frobenius norm of $K'$ leads to $s_\lambda$, and $K'$ has the largest eigenvalue $((N-1)\kappa_2'+\kappa_1')$ for arbitrary $N$. The eigenspace of $\Theta$ corresponding to $\lambda_{max}$ is also the same as $F^*$. It is spanned by eigenvectors $\nu_k$ ($k=1,...,C$).

\section{Eigenvalue statistics of $A$}
\renewcommand{\theequation}{D.\arabic{equation} }
\setcounter{equation}{0}
\subsection{Derivation of Theorem 5.1}
The metric tensor $A_k$ can be represented by $A_k = \nabla_h f_k \nabla_h f_k^{\top}/N$, where $\nabla_h f_k$ is an $M_h \times N$ matrix and its columns are the gradients on each input, i.e., $\nabla_{h} f_k(n)$ $(n=1,...,N)$. Let us introduce the $N \times N$ dual matrix of $A_k$, i.e., $A^*_k :=\nabla_h f_k^{\top}\nabla_h f_k/N$. It has the same non-zero eigenvalues as $A_k$ by definition. Its $nm$-th entry is given by 
\begin{align}
 (A^{*}_k)_{nm} &= \nabla_{h} f_k(n)^{\top} \nabla_{h} f_k(m)/N \nonumber \\ 
 &= \sum_{l=0}^{L-1} \sum_{i,j,j'} W^{l+1}_{ji} W^{l+1}_{j'i} \delta_j^{l+1}(n) \delta_{j'}^{l+1}(m)/N \nonumber \\
 &\sim \sigma_w^2 \sum_{l=1}^{L} \sum_{j} \delta_j^{l}(n) \delta_j^{l}(m)/N,
\end{align}
in the large $M$ limit. Accordingly, we have
\begin{align}
\begin{aligned}
A^{*}_k &= \frac{\tilde{\alpha}}{N} \bar{A}^{*} + \frac{1}{N}o(1), 
\\ (\bar{A}^{*})_{nm} &:= \tilde{\kappa}_1 \ \ (n=m), \ \ \tilde{\kappa}_2 \ \ (n \neq m).
\end{aligned}
\end{align}
$\tilde{\kappa}_1$ is positive by definition, and $\tilde{\kappa_2}$ is positive under the condition of the activation functions (ii). 

The eigenvalue statistics are easily derived from the leading term $\bar{A}^{*}$. We can derive the mean of the eigenvalues as $m_{\lambda} \sim \mathrm{Trace}(\frac{\tilde{\alpha}}{N}\bar{A}^{*} )/M_h$ and the second moment as $s_{\lambda} \sim ||\frac{\tilde{\alpha}}{N}\bar{A}^{*}||_{F}^2/M_h$, where $M_h=\tilde{\alpha}M$.  We can determine the largest eigenvalue because we explicitly obtain the eigenvalues of $\bar{A}^{*}$; $\lambda_{1}= (N-1)\tilde{\kappa}_2+\tilde{\kappa}_1$ with an eigenvector $\tilde{\nu}:=(1,...,1)$ and $\lambda_i=\tilde{\kappa}_1 -\tilde{\kappa}_2$ with eigenvectors $e_1 -e_i$ ($i=2,...,N$). The vector $e_i$ denotes a unit vector whose entries are $1$ for the $i$-th entry and $0$ otherwise. The largest eigenvalue is given by $\lambda_{1}$. 

The eigenvector of $A_k$ corresponding to $\lambda_{max}$ is constructed from $\tilde{\nu}$. Let us denote by $v$ an eigenvector of $A_k$ satisfying $A_kv=\lambda_{max}v$. Multiplying both sides by $\nabla_h f_k^{\top}$, we get
\begin{equation}
A^*_k(\nabla_h f_k^{\top}v)= \lambda_{max}\cdot(\nabla_h f_k^{\top}v).
\end{equation}
This means that $\nabla_h f_k^{\top}v$ is the eigenvector of $A_k^*$ and equals $\tilde{\nu}$. Multiplying both sides of $\tilde{\nu}=\nabla_h f_k^{\top}v$ by $ \frac{1}{N}\nabla_h f_k$, we get 
\begin{equation}
 \mathrm{E}[\nabla_h f_k]=A_k v, 
\end{equation}
which equals $ \lambda_{max}v$ by definition of $v$. 
As a result, we obtain $v= \mathrm{E}[\nabla_h f_k]$ up to a scale  factor.

We can also evaluate eigenvalue statistics of $A=\sum_k^C A_k$ in analogy with $F$. Note that we can represent $A$ as $A = \tilde{R}\tilde{R}^{\top}$ with
\begin{eqnarray}
\tilde{R} &:=& 
\frac{1}{\sqrt{N}}[\nabla_h f_1 \ \ \nabla_h f_2 \ \ \cdots \ \ \nabla_h f_C]. 
\end{eqnarray}
Its columns are the gradients on each input, i.e., $\nabla_{h} f_k(n)$ $(n=1,...,N)$. We can introduce a dual matrix  $\tilde{R}^{\top}\tilde{R}$ and obtain the following eigenvalue statistics of $A$:
\begin{align}
\begin{aligned}
m_{\lambda} &\sim \tilde{\kappa}_1 \frac{C}{M}, \ \ s_{\lambda} \sim \tilde{\alpha} \left ( \frac{N-1}{N} \tilde{\kappa}_2^2 + \frac{\tilde{\kappa}_1^2 }{N} \right )C, \\ 
\lambda_{max} &\sim \tilde{\alpha} \left(\frac{N-1}{N} \tilde{\kappa}_2 + \frac{\tilde{\kappa}_1}{N}\right). \label{eqe6:0721}
\end{aligned}
\end{align}

\subsection{Diagonal blocks}

In the same way as in the FIM, we can also evaluate the eigenvalue statistics of the diagonal blocks $A^{ll}_k$ ($l=0,...,L-1$).  
The metric tensor $A^{ll}_k$ can be represented by $A^{ll}_k = \nabla_{h^l} f_k \nabla_{h^l} f_k^{\top}/N$. Consider its dual, i.e., $A^{ll*}_k =\nabla_{h^l} f_k^{\top} \nabla_{h^l} f_k/N$. 
Note that we partitioned $A_k$ into $L^2$ block matrices whose $(l,l')$-th block is expressed by an $M_l \times M_{l'}$ matrix:
\begin{equation}
 A_k^{ll'} :=(W^{l+1})^\top \mathrm{E} [\delta^{l+1}_k (\delta_k^{l'+1})^{\top} ]W^{l'+1}. \label{eq0407:13}
\end{equation}
In the large $M$ limit, we have asymptotically 
\begin{align}
\begin{aligned}
A^{ll*}_k &= \frac{1}{N} \bar{A}^{ll*} + \frac{1}{N}o(1), \\ 
(\bar{A}^{ll*})_{nm}&:= \sigma_w^2 \tilde{q}^{l+1}_{{1}} \ \ (n=m), \ \ \sigma_w^2\tilde{q}^{l+1}_{{2}} \ \ (n \neq m). 
\end{aligned}
\end{align}
The eigenvalue statistics of $A^{ll}_k$ are asymptotically evaluated as 
\begin{equation}
m_{\lambda} \sim \sigma_w^2 \frac{\tilde{q}^{l+1}_{{1}}}{M_l}, \ \ s_{\lambda} \sim \sigma_w^4\left ( \frac{N-1}{N} (\tilde{q}_{{2}}^{l+1})^2 + \frac{(\tilde{q}^{l+1}_{{1}})^2 }{N} \right ), \ \
\lambda_{max} \sim \sigma_w^2\left(\frac{N-1}{N} \tilde{q}^{l+1}_{{2}} + \frac{\tilde{q}^{l+1}_{{1}}}{N}\right). \nonumber
\end{equation}
The eigenvector of $A_k^{ll}$ corresponding to $\lambda_{max}$ is $\mathrm{E}[\nabla_ {h^l} f_k].$
 We can also derive the eigenvalue statistics of the summation $A^{ll}=\sum_k^C A_k^{ll}$; the mean and second moment are multiplied by $C$.

%\section*{Acknowledgment}
%This work was partly supported by a Grant-in-Aid for Young Scientists (19K20366) from the Japan Society for the Promotion of Science (JSPS).

\bibliographystyle{unsrtnat}
\bibliography{sample2}

\end{document}